\documentclass[]{joyenfra}

\usepackage[toc,page,header]{appendix}

\usepackage{minitoc}
\usepackage{cleveref} 
\usepackage{subcaption}
\usepackage{booktabs}
\usepackage{graphicx}
\usepackage{subcaption}

\newcolumntype{L}[1]{>{\raggedright\arraybackslash}p{#1}}
\newcolumntype{C}[1]{>{\centering\arraybackslash}p{#1}}
\newcolumntype{Y}{>{\raggedright\arraybackslash}X}

\usepackage{fancyhdr} 

\usepackage{natbib}
\usepackage{latexsym}

\usepackage{url}
\usepackage{amssymb}
\usepackage[utf8]{inputenc}
\usepackage{microtype}
\usepackage{booktabs}
\usepackage{pifont} 
\usepackage{multirow}
\usepackage{makecell}
\usepackage{paralist}
\usepackage{xspace}
\usepackage{color}
\usepackage{xcolor}
\usepackage{colortbl}
\usepackage{adjustbox}
\usepackage{hyperref} 
\usepackage[edges]{forest}
\usepackage{tikz} 
\usepackage{caption}
\usepackage{amsfonts}

\hypersetup{
    colorlinks,
    linkcolor={blue!80!black},
    citecolor={blue!80!black},
}
\tikzset{
    root/.style =             {align=center, text width=1cm, rounded corners=3pt, line width=0.3mm, fill=gray!10, draw=gray!80, font=\small},
    demographic/.style =         {align=center, text width=1.8cm, rounded corners=3pt, line width=0.3mm, fill=blue!10, draw=blue!80, font=\footnotesize},
    demographic_work/.style =    {align=center, text width=10cm, rounded corners=3pt, line width=0.3mm, fill=blue!10, draw=blue!0, font=\footnotesize},
    character/.style =         {align=center, text width=1.8cm, rounded corners=3pt, line width=0.3mm, fill=red!10, draw=red!80, font=\footnotesize},
    character_work/.style =    {align=center, text width=10cm, rounded corners=3pt, line width=0.3mm, fill=red!10, draw=red!0, font=\footnotesize},
    personalization/.style =           {align=center, text width=1.8cm, rounded corners=3pt, line width=0.3mm, fill=cyan!10, draw=cyan!80, font=\footnotesize},
    personalization_work/.style =      {align=center, text width=10cm, rounded corners=3pt, line width=0.3mm, fill=cyan!10, draw=cyan!0, font=\footnotesize},
    risk/.style =         {align=center, text width=1.8cm, rounded corners=3pt, line width=0.3mm, fill=orange!10, draw=orange!80, font=\footnotesize},
    risk_work/.style =    {align=center, text width=10cm, rounded corners=3pt, line width=0.3mm, fill=orange!10, draw=orange!0, font=\footnotesize},
}

\usepackage{CJK}

\title{Thousand-GPU Large-Scale  Training and Optimization Recipe for AI-Native Cloud Embodied Intelligence Infrastructure}

\affiliation[1]{AI Infra Team at JDT\quad $^2$Tsinghua University\quad $^3$Peking University\quad  $^4$Tianjin University
\quad $^5$Beihang University
\quad $^6$University of Science and Technology of China}

\contribution{Full author list in Contributions}

\abstract{
Embodied intelligence is a key step towards Artificial General Intelligence (AGI), yet its development faces multiple challenges including data, frameworks, infrastructure, and evaluation systems. To address these issues, we have, for the first time in the industry, launched a cloud-based, thousand-GPU distributed training platform for embodied intelligence, built upon the widely adopted LeRobot framework, and have systematically overcome bottlenecks across the entire pipeline. At the data layer, we have restructured the data pipeline to optimize the flow of embodied training data. In terms of training, for the GR00T-N1.5 model, utilizing thousand-GPU clusters and data at the scale of hundreds of millions, the single-round training time has been reduced from 15 hours to just 22 minutes, achieving a 40-fold speedup. At the model layer, by combining variable-length FlashAttention and Data Packing, we have moved from sample redundancy to sequence integration, resulting in a 188\% speed increase; $\pi_{0.5}$  Attention optimization has accelerated training by 165\%; and FP8 quantization has delivered a 140\% speedup. On the infrastructure side, relying on high-performance storage, a 3.2T RDMA network, and a Ray-driven elastic AI data lake, we have achieved deep synergy among data, storage, communication, and computation. At the training level, we propose and implement a fully asynchronous strategy training pipeline for the first time — RL-VLA$^{3}$, a triple-level asynchronous architecture covering the entire process from environment interaction and trajectory generation to policy network updates, achieving a maximum throughput increase of 126.67\%. We have also built an end-to-end evaluation system, creating a closed loop from training to simulation to assessment. This framework has already been fully validated on thousand-GPU clusters, laying a crucial technical foundation for the development and application of next-generation autonomous intelligent robots, and is expected to accelerate the arrival of the era of human-machine integration.
}

\date{\today}
\correspondence{Junwu Xiong at \email{xiongjunwu.1@jd.com}}

\begin{document}

\maketitle


\section{Introduction}

\begin{figure}
    \centering
    \includegraphics[width=1\linewidth]{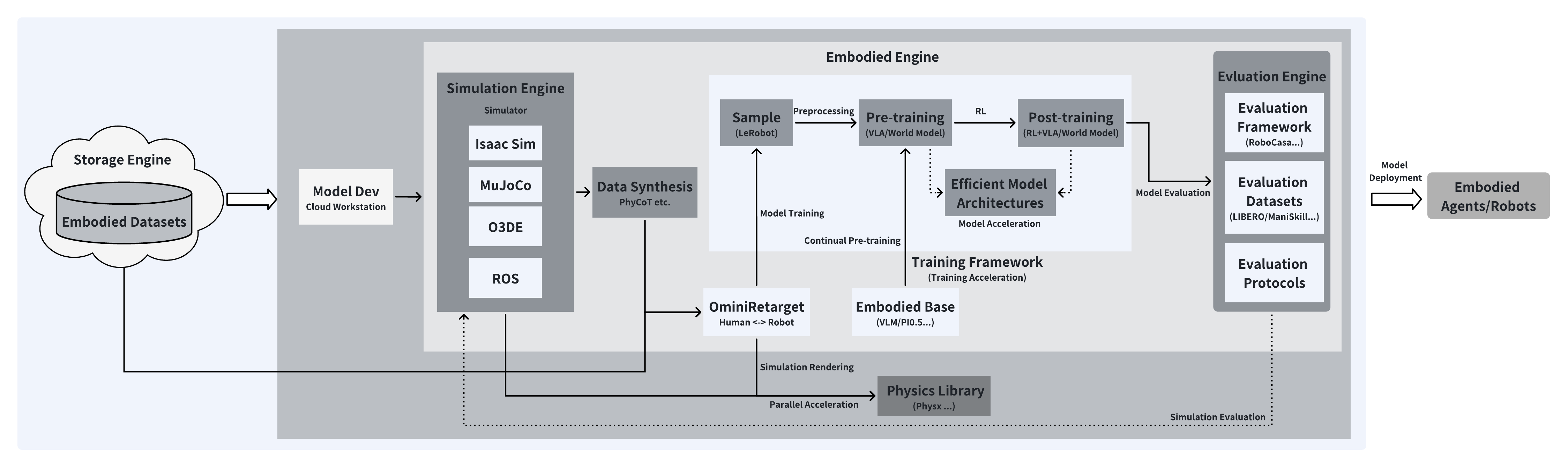}
    \caption{AI-Native Cloud Embodied Intelligence Infrastructure }
    \label{fig:ai-native}
\end{figure}
Designing agents with physical-world action capabilities that reach or surpass human-level is one of the key technical paths toward Artificial General Intelligence (AGI) \cite{lecun2022path}, a direction known as ``Embodied AI.'' Embodied intelligence is currently experiencing a period of market expansion: Driven by global innovation and market demand, the embodied AI industry is expanding rapidly at an annual growth rate achieving 39\% \cite{marketsandmarkets2024embodied}, with the market size expected to surpass $\$100$ billion by 2035 \cite{gaosheng}.

In the field of embodied intelligence, traditional dynamics-based embodied strategies mainly address pre-defined task sets in controlled environments, but they have weak generalization abilities and are sensitive to noise in real-world scenarios \cite{tutorial:capuano2025robot}.
In recent years, breakthroughs in multimodal understanding, reasoning, and utilization of vision and language foundation models \cite{VideoVLA:shenvideovla} have propelled the rapid development of Vision-Language-Action (VLA) models. 
VLA models, with powerful cross-modal sequential modeling capabilities, integrate perception, understanding, reasoning, and action generation end-to-end, enabling direct mapping from visual observation and language instructions to physical actions. 
Existing research has demonstrated their outstanding versatility, flexibility, and generalization to complex environments \cite{pi0.5:zhou2025vision,black2410pi0,bjorck2025gr00t}. 
Simultaneously, the embodied intelligence ecosystem continues to integrate with world models~\cite{world-model:hafner2023mastering}, robotic hardware~\cite{aloha:fu2024mobile}, open-source datasets~\cite{cadene2024lerobot}, simulation environments~\cite{NVIDIA_Isaac_Sim,zakka2025mujoco}, training algorithms~\cite{schulman2018highdimensionalcontinuouscontrolusing}, and computational resource\cite{computation:zitkovich2023rt} supply, collectively forming the AI-Native Embodied Intelligence Infrastructure (as shown in Figure~\ref{fig:ai-native}), which provides a solid foundation for the continuous optimization and iteration of VLA-based embodied intelligence systems.

Success stories such as the game of Go \cite{alphazero} and LLMs~\cite{GPT4:achiam2023gpt} show that large-scale, general-purpose computation is essential to move embodied intelligence from scientific research to industrial application. However, achieving large-scale training for embodied intelligence models poses higher demands and challenges for AI infrastructure.

\textbf{Training framework challenges:} For thousand-GPU scale embodied training, there is a lack of industrial-grade systems that seamlessly connect simulation, training, and evaluation; multi-dimensional parallelism is complex, making communication and load balancing difficult; large batch data loading is easily blocked by I/O, leading to unstable training and low compute utilization, which hinders efficient large-scale training.

\textbf{Data engine challenges:} Mixed storage of multimodal files increases system complexity, and high concurrency puts pressure on metadata processing; frequent file operations lead to increased latency and reduced throughput, affecting GPU data supply; uneven load among nodes may block distributed training; the complexity of data preprocessing further burdens storage. Traditional data lakes struggle to dynamically allocate large files, and serial processing can lead to resource idleness and task blockage, lacking elastic scalability and failing to meet cloud-native high-availability requirements.

\textbf{Model computation challenges:} Insufficient dynamic computation and memory optimization, padding in traditional attention mechanisms generates invalid tokens, causing wasted compute and memory redundancy; inefficient training data organization leads to resource waste and low hardware utilization due to short data being padded to fixed lengths; model inference and edge deployment are constrained by real-time requirements and compute resources, making efficient compression and acceleration of small-parameter models key for deployment.

Based on these challenges, the AI Infra team at JDT has proposed and implemented a thousand-GPU distributed training framework for embodied intelligence based on the JD Cloud JoyBuilder platform, using the open-source LeRobot framework as its foundation, as shown in Figure~\ref{fig:joyEnfra}. It is equipped with a 3.2T RDMA backend network supporting up to ten thousand GPUs and a flexible VPC frontend network, systematically overcoming bottlenecks across the entire pipeline. Through Yunhai high-performance storage and a Ray-driven elastic AI data lake, the training data pipeline is collaboratively optimized to meet embodied model training needs. For the GR00T-N1.5~\cite{bjorck2025gr00t} model, in thousand-GPU clusters and billion-scale data scenarios, single-round training time is reduced from 15 hours to 22 minutes, a 40-fold speedup; combining variable-length FlashAttention and Data Packing achieves sequence integration from sample redundancy, accelerating by 188\%; custom $\pi_{0.5}$ architecture and post-training quantization improve training and inference efficiency by 165\%; FP8 quantization accelerates by 140\%; and an end-to-end evaluation system is built to connect the training-simulation-evaluation pipeline. At the VLA training level, we propose and implement a fully asynchronous strategy training pipeline for the first time — RL-VLA$^3$~\cite{guan2026rl}, a triple-level asynchronous architecture covering the entire process from environment interaction and trajectory generation to policy network updates. On the LIBERO benchmark, this framework achieves a maximum throughput improvement of 59.25\% compared to existing synchronous training strategies. After further optimization through a decoupling strategy, the throughput improvement reaches 126.67\%. This framework has been validated at scale on 256 GPUs clusters, providing robust infrastructure support for the industrialization of embodied intelligence.


\section{Core Architecture and Model Optimization }

\subsection{Overall Architecture Design}
\begin{figure}[h!]
    \centering
    \includegraphics[width=1\linewidth]{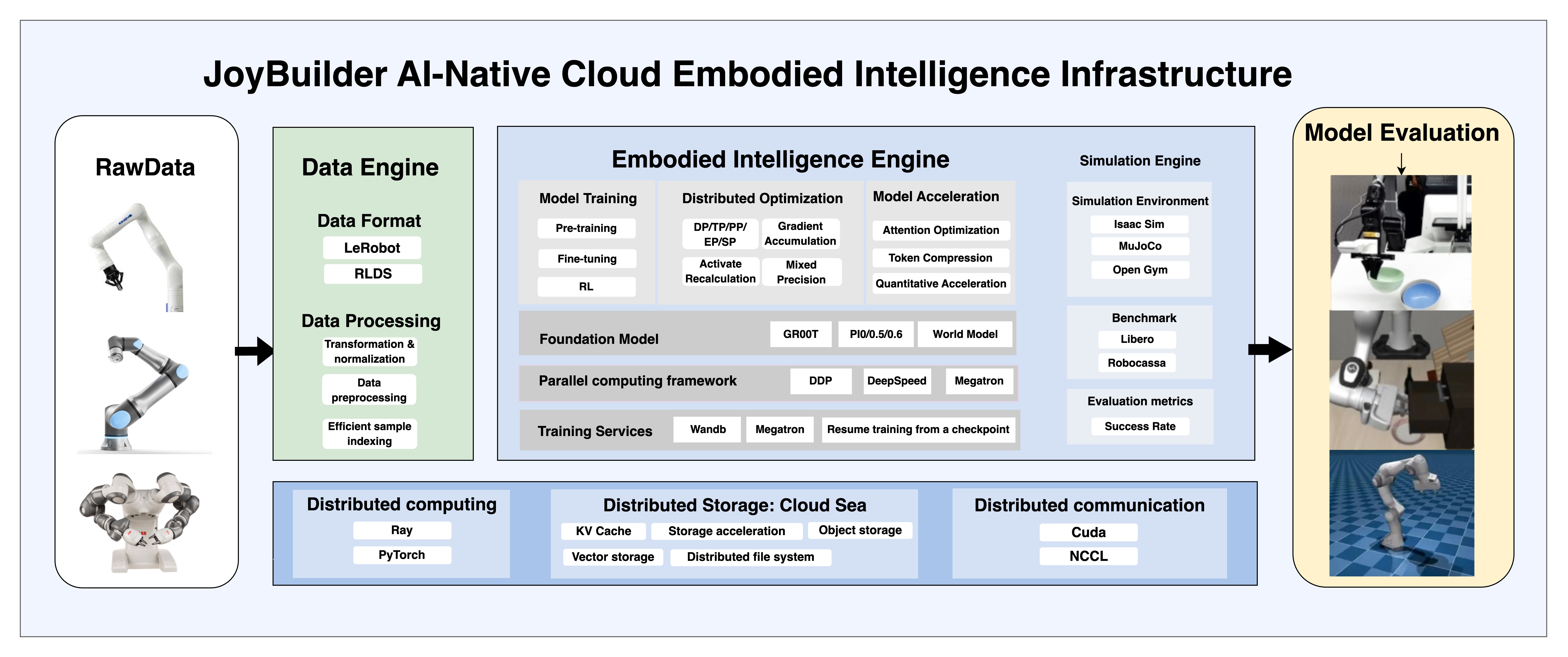}
    \caption{The overview of Joybuilder AI-Native Cloud Embodied Intelligence Infrastructure }
    \label{fig:joyEnfra}
\end{figure}

With the continuous growth in the scale of embodied intelligence models and the complexity of data, efficient distributed frameworks are playing an increasingly important role in large-scale training and multi-scenario adaptation. These frameworks integrate VLA models~\cite{kim2024openvla}, standardized data formats, and unified simulation interfaces, significantly lowering the R\&D threshold. Mainstream open-source frameworks such as Lerobot~\cite{cadene2024lerobot} and RLinf~\cite{yu2025rlinf,zang2025rlinf} have their respective strengths in model support, data adaptation, and simulation environments, jointly promoting the learning of generalized skills for robots. LeRobot~\cite{cadene2024lerobot} is renowned for its strong usability and active community, natively supporting standard data formats and various simulation environments, and is convenient for real robot integration, making it suitable for rapid prototyping and academic exploration. RLinf focuses on reinforcement learning and high-fidelity simulation, offering flexible sampling but requiring more customization. SimpleVLA-RL~\cite{li2025simplevla} and Dexbotic~\cite{xie2025dexbotic} focus on minimal packaging and industrial data pipelines, but have limited functionality and community support. Overall, LeRobot’s usability and ecosystem make it an ideal framework for research and development. Based on the LeRobot framework, we have integrated NVIDIA’s high-fidelity simulation platform to build a new generation of cloud-native embodied intelligence training framework, as shown in Figure~\ref{fig:joyEnfra}, accelerating the development of embodied intelligence.

\subsubsection{LeRobot and Nvidia Open-Source Ecosystem}
LeRobot is a lightweight embodied intelligence training framework launched by the Hugging Face community, supporting standard Dataset formats and various simulation environments, and can be easily integrated with real robots, making it suitable for rapid experimentation and algorithm verification. However, when faced with high-fidelity simulation and large-scale training requirements, a single open-source framework is still insufficient. NVIDIA has built a full-stack ecosystem covering simulation, training, and deployment, providing high-fidelity simulation platforms such as Isaac Lab~\cite{mittal2025isaac} and Omniverse, optimizing the training process, and supporting efficient migration from simulation to real-world with Isaac ROS and Jetson hardware. This ecosystem performs excellently in complex tasks but requires higher hardware and engineering capabilities from developers, forming a complementary relationship with community frameworks.

\subsubsection{Cloud-Native Cloud Embodied Intelligence Training Infrastructure}

We have integrated NVIDIA high-fidelity simulation and LeRobot standardized data to build a new generation of cloud-native embodied intelligence training framework, systematically integrating data processing, model training, simulation evaluation, and distributed computing power, providing full-stack support for large-scale training and evaluation of VLA models. The framework consists of four layers: the data layer is compatible with mainstream formats (such as LeRobot, RLDS), supports efficient preprocessing and streaming loading, meeting the needs of massive samples; the training layer supports pre-training, fine-tuning, and reinforcement learning, integrates distributed tools such as PyTorch DDP and DeepSpeed~\cite{yao2023deepspeed}, improves cluster compute utilization, and offers experiment tracking and checkpoint recovery; the simulation evaluation layer uniformly connects to multiple environments (Open Gym~\cite{brockman2016openai}, Mujoco~\cite{zakka2025mujoco}, Isaac Sim~\cite{NVIDIA_Isaac_Sim}), with built-in automated evaluation processes for quantitative model validation; the distributed infrastructure relies on CUDA, NCCL, and Ray to achieve efficient communication, storage acceleration, and resource scheduling, fully adapting to cloud-native environments. The overall framework combines open-source usability with industrial-grade system design, facilitating the rapid deployment of embodied intelligence technologies.

This framework outlines a complete technology stack from data processing to resource scheduling. To support training of thousand-GPU clusters and models with tens or even hundreds of billions of parameters, the key lies in efficiently and flexibly organizing and scheduling large-scale computing resources---this is the core mission of distributed parallel technology. Next, we will delve into the design and practice of distributed parallel strategies from an architectural perspective, specifically analyzing how multidimensional parallel combinations can achieve optimal resource allocation and maximize training efficiency in real-world model training scenarios.

\subsubsection{Distributed Parallel Training}
\begin{figure}
    \centering
    \includegraphics[width=0.8\linewidth]{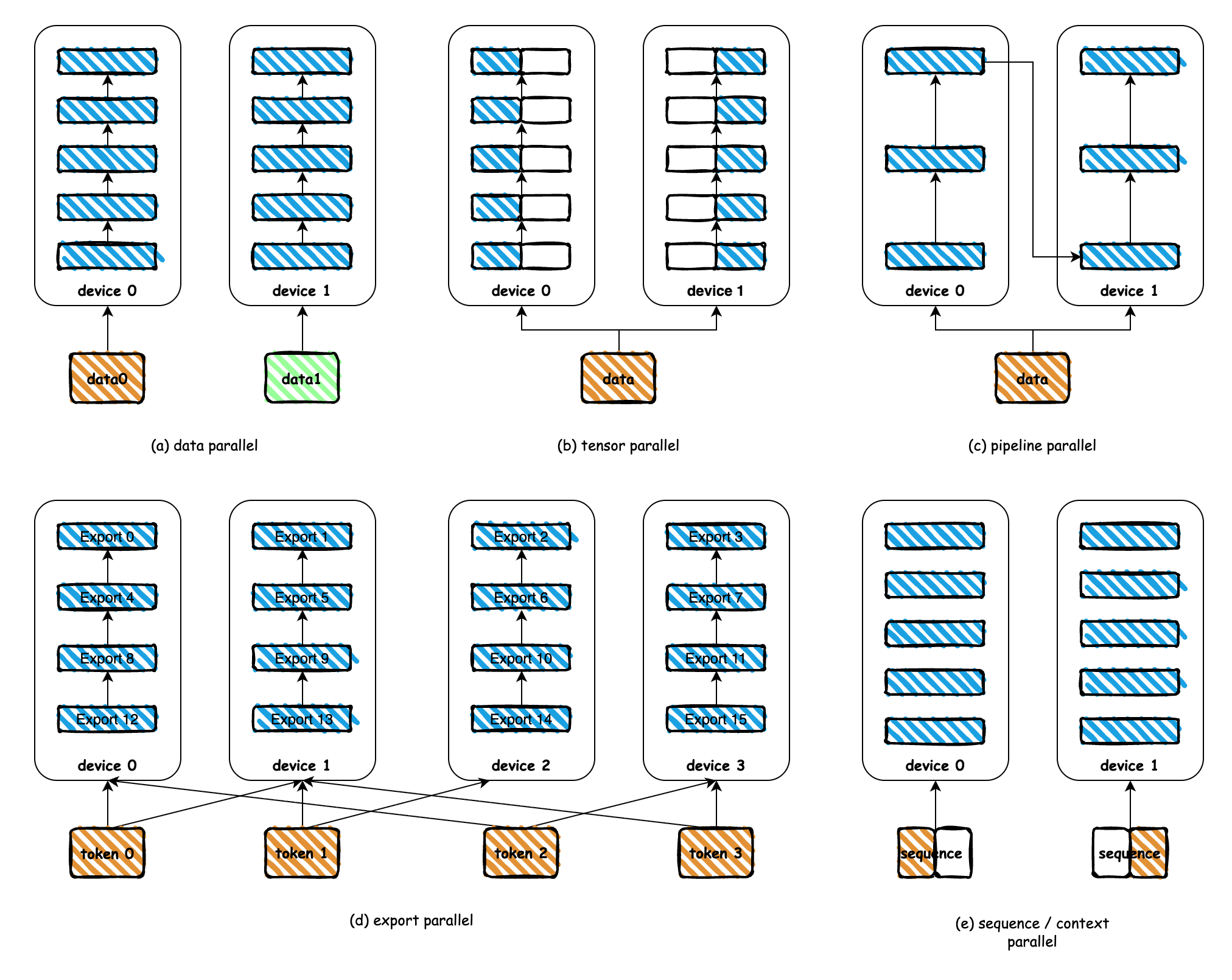}
    \caption{Schematic diagram of distributed training parallel strategies}
    \label{fig:4d}
\end{figure}

In actual training of models with tens or hundreds of billions of parameters, a single parallel strategy is often insufficient, requiring multidimensional combinations:
\begin{itemize}
    \item \textbf{3D Parallelism:} Data parallelism + pipeline parallel parallelism + tensor parallelism. Tensor parallelism decomposes large layers within nodes (high-speed interconnect), pipeline parallelism decomposes model depth across nodes, and data parallelism replicates the above mixed parallel groups to handle larger batches.
    \item \textbf{Expert Parallel : }On top of 3D parallelism, for MoE models, expert parallelism is introduced. Expert parallel groups can be nested or separated from tensor parallel groups to optimize communication.
    \item \textbf{Sequence Parallelism:} If sequence length leads to activation memory explosion, sequence parallelism is further introduced, usually closely combined with tensor parallelism.
\end{itemize}

This multidimensional, hierarchical parallel strategy design is the core of modern large model training systems, requiring trade-offs between computational efficiency, memory usage, and communication overhead.

\begin{table}[htbp]
\centering
\small
\caption{Comparison of Distributed Training Parallelism Strategies:Data Parallelism (DP), Pipeline Parallelism (PP), Tensor Parallelism (TP), Expert Parallelism (EP) and Sequence Parallelism (SP). }
\begin{tabular}
{@{}cL{3.4cm}L{3.0cm}L{3.0cm}L{3.8cm}@{}}
\toprule
\textbf{Strategy} & \textbf{Partitioned Object} & \textbf{Communication Pattern} & \textbf{Key Problem Solved} & \textbf{Scalability Limitation} \\ \midrule
DP & Training data & All-Reduce & Data throughput, compute acceleration & Per-device memory, global batch size \\
PP & Model layers & Point-to-point (adjacent stages) & Model depth (layers) & Pipeline bubbles, stage load balancing \\
TP & Intra-layer weight matrices & All-Reduce / All-Gather (within layer) & Model width (layer size) & Frequent intra-layer communication, bandwidth \& latency sensitive \\
EP & Expert sub-networks & All-to-All (token routing) & Total parameter count & Load balancing, token-routing communication overhead \\
SP & Activation tensors (sequence dim) & All-Gather / Reduce-Scatter & Activation memory for long sequences & Communication overhead of sequence operations \\ \bottomrule
\end{tabular}

\end{table}

\paragraph{Distributed Data Parallel (DDP)} is a distributed training paradigm based on data parallelism. Its core goal is to efficiently train large models on single-node or multi-node clusters with multiple computing devices (such as multiple GPUs) by parallelizing data loading and gradient synchronization, achieving near-linear training acceleration. PyTorch DDP maximizes distributed compute resource utilization through three core technologies: a multi-process architecture, peer-to-peer All-Reduce communication, and bucketed gradient synchronization, while maintaining model replica consistency. The core workflow is as follows:
\begin{enumerate}
    \item \textbf{Data Sharding:} Each process loads a different shard of the global dataset, generating a local mini-batch;
    \item \textbf{Forward Pass:} The local model replica performs forward computation on the shard data;
    \item \textbf{Backward Pass:} Local gradients are computed based on the loss;
    \item \textbf{Gradient Synchronization:} AllReduce collective communication is triggered, summing and averaging the gradients of the same parameters across all processes, synchronizing them to each process;
    \item \textbf{Parameter Update:} Each process independently updates its local model parameters with the synchronized gradients.
\end{enumerate}

Through DDP-based distributed training, we have implemented GR00T and LeRobot training on thousand-GPU clusters.

\begin{figure}[h!]
    \centering
    \includegraphics[width=0.8\linewidth]{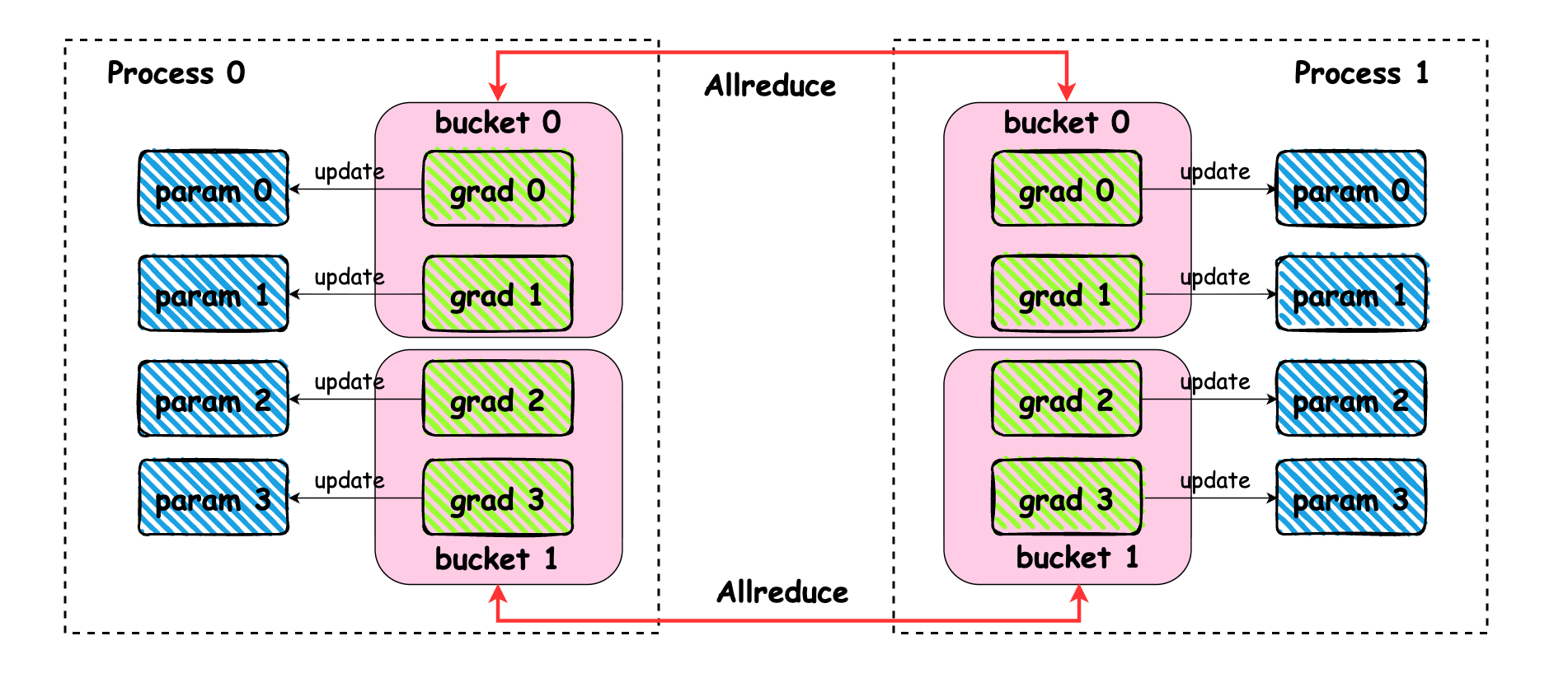}
    \caption{Diagram illustrating Distributed Data Parallelism (DDP)}
    \label{fig:DDP}
\end{figure}


\subsection{Model-Level Training Optimization for Embodied Models}

The cloud-native embodied distributed framework provides an efficient foundation for embodied model training, while computational optimization at the model layer is key to further unleashing compute potential. In cloud-native embodied models, VLA is responsible for mapping multimodal perception and high-level instructions to action sequences; the World Model realizes safer and more efficient autonomous decision-making through environment modeling and simulation reasoning. The two together form the decision-making core.
\begin{figure}[h!]
    \centering
    \includegraphics[width=0.8\linewidth]{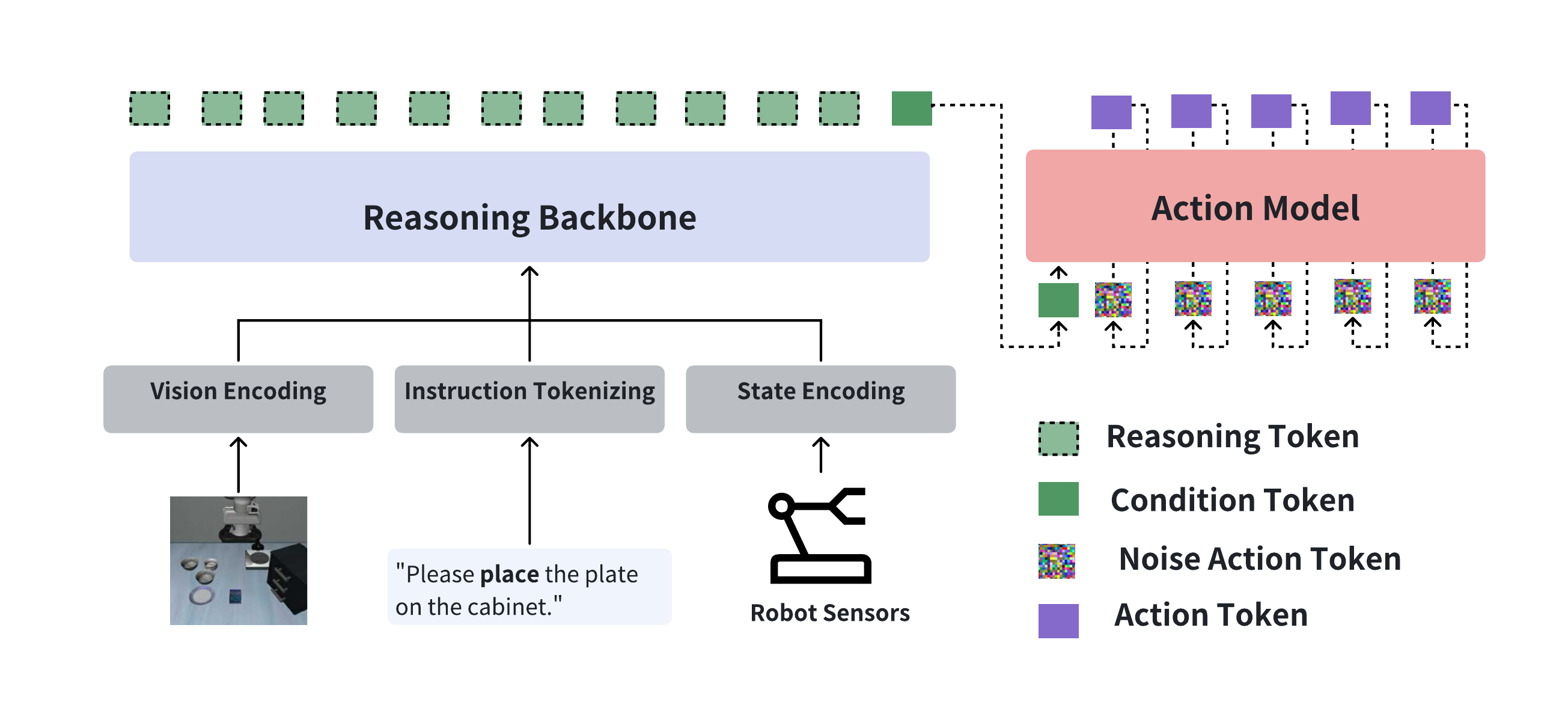}
    \caption{Schematic diagram of a typical VLA model architecture}
    \label{fig:vla}
\end{figure}

Embodied base models represented by VLA have undergone technical evolution from basic capability exploration to efficient, scenario-oriented multimodal fusion architectures. Early models like PaLM-E validated the underlying ``vision-language-action'' linkage, while models like GR00T and $\pi$ series continuously enhance multi-task adaptation and cross-scenario action generalization. In 2025, VLA entered a period of intensive efficient iteration, breaking deployment bottlenecks through lightweight design and structural optimization, and expanding action scenarios with reinforcement learning, becoming a core requirement for large-scale cloud applications.

Although current VLA models have distinctive architectures, they generally follow the universal computational paradigm of ``multimodal encoding---large model decision-making---action generation'': visual observations, text instructions, and embodiment states are encoded by encoders, input into LLMs for planning and decision-making, and then output as continuous action sequences by the action generation module. The $\pi$ series ($\pi_0$~\cite{black2410pi0}, $\pi_{0.5}$~\cite{pi0.5:zhou2025vision}) and GR00T N1.5~\cite{bjorck2025gr00t} are typical representatives of this paradigm. $\pi_{0.5}$ improves the generalization ability of robotic arms in home environments through hierarchical reasoning and multi-source data training; GR00T N1.5 achieves efficient collaboration and task adaptability with a tightly coupled dual-system architecture and innovative data pyramid strategy, and realizes integrated data, training, and simulation through the NVIDIA Isaac ecosystem~\cite{NVIDIA_Isaac_Sim}. Based on the cloud-native embodied training framework and typical models, the team has optimized distributed training for GR00T N1.5 at a thousand-GPU scale, significantly improving large-scale training performance. At the same time, efficiency improvements have also been made for the $\pi$ series models, mainly in three aspects:
\begin{enumerate}
    \item Optimizing the attention structure for long-sequence processing with Efficient Attention or Mamba Block to reduce computational complexity;
    \item Filtering image and text tokens to reduce redundant data and ineffective computation;
    \item Compressing model size using quantization techniques to accelerate computation while maintaining model accuracy.
\end{enumerate}

\subsubsection{Attention Dynamic Computation and Memory Optimization}

\paragraph{(1) Variable-Length Flash-Attention: Eliminating Padding Compute Waste}
As the ``understanding core'' of VLA, VLM undertakes cross-modal understanding of images and text, and its computational efficiency directly determines the training and inference performance of the entire VLA model. During model training and inference, multimodal inputs (such as image patch tokens and text sequences) naturally have inconsistent lengths. The traditional approach uses padding to fill all inputs to a fixed length, causing many invalid padding tokens to participate in attention computation, resulting in wasted compute and memory redundancy.

\begin{figure}[h!]
    \centering
    \includegraphics[width=1\linewidth]{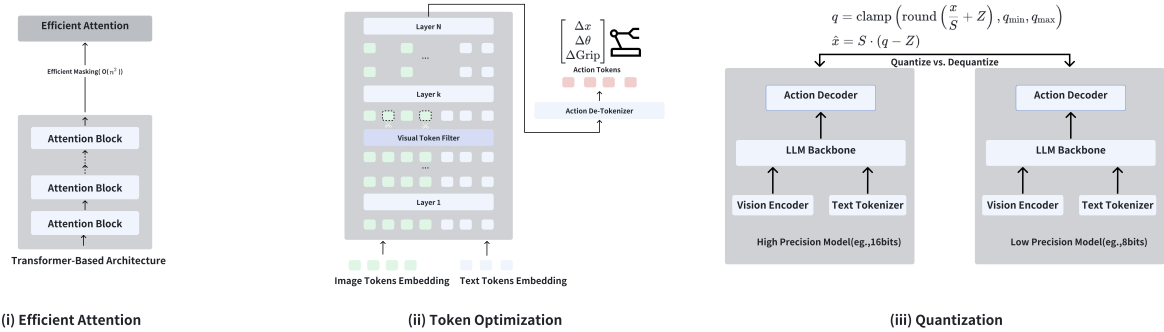}
    \caption{Key optimization strategies for the VLA model}
    \label{fig:vla_policy}
\end{figure}

This work focuses on the attention computation of the Qwen2.5-VL model LM Decoder, systematically analyzing the current attention paradigms supported by the official Transformers library and introducing data packing strategies to further compress redundant computation. Specifically, the visual-side attention module, when Flash-Attention-2 is enabled, directly calls its variable-length (varlen) interface to perform matrix operations only on valid visual tokens; on the language model side, visual features and text embeddings are concatenated into a mixed sequence and then fed into the LM Decoder for autoregressive modeling. When Flash-Attention-2 is effective, the LM Decoder dynamically selects the ``variable-length'' or ``non-variable-length'' computation path according to the sequence distribution, achieving optimal matching of computational intensity and memory bandwidth.

Based on Flash-Attention, we optimize for variable-length computation, calculating only on valid token sequences, greatly improving model training and inference efficiency while maintaining stable model performance:
\begin{itemize}
    \item As sequence length or padding rate increases, the time savings of variable-length computation become more significant;
    \item As sequence length increases, the TFLOPS of variable-length attention gradually approaches that of non-variable-length, and with moderate padding rates, can even surpass non-variable-length;
    \item Variable-length attention maintains stable leaderboard performance compared to non-variable-length.
\end{itemize}

\paragraph{(2) Data Packing: From Sample Redundancy to Sequence Integration}
In large model training, training data often consists of text sequences of varying lengths. During data preprocessing, the data packing strategy is used to pack different samples into one sequence, so that the data received during training is in a no-padding format, and \texttt{flash\_attn} is called for attention computation. The traditional approach fills texts shorter than the fixed length with special tokens, resulting in computational waste and inefficient training.

To address the efficiency issue caused by inconsistent data lengths in large model training, this study innovatively combines the Data Packing strategy with Flash-Attention. Our proposed optimization intelligently concatenates multiple shorter training samples to construct a sequence close to the model's maximum context length, thereby minimizing or even eliminating the use of padding tokens. This sequence integration method provides a new technical path for efficient training of large-scale language models, especially suitable for handling multimodal training data with significant length differences. Specific experimental results are shown in section~\ref{expr:datapacking}.

\begin{figure}[h!]
    \centering
    \includegraphics[width=0.8\linewidth]{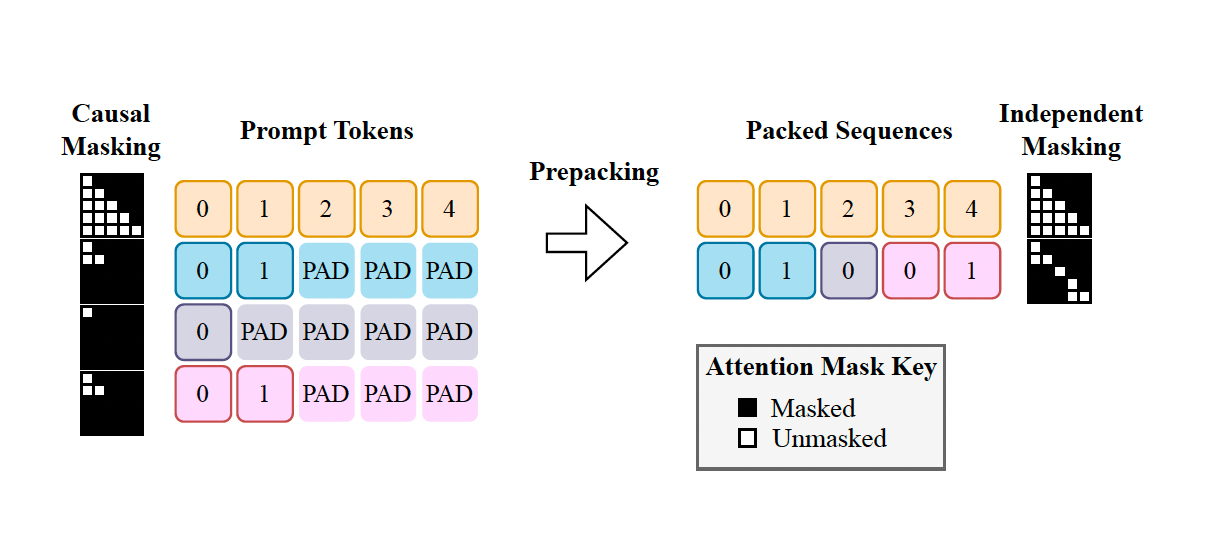}
    \caption{Padding and packing processing~\cite{zhao2024prepacking}}
    \label{fig:placeholder}
\end{figure}

\subsubsection{Acceleration Optimization for VLA Model Architecture: $\pi_{0.5}$ Case Study}

To improve the training efficiency of the $\pi_{0.5}$ VLA model, this work proposes a customized optimization strategy targeting redundant computation and memory overhead in its attention mechanism. The strategy focuses on the waste of computational resources caused by invalid tokens during the processing of multimodal inputs (images, text, states, and actions), especially redundant information in visual and textual modalities.

Specifically, the $\pi_{0.5}$ model uses a Vision-Language Model (VLM) to process image, text, and state information during training, and generates action outputs through an independent action expert module. In this process, the attention mechanism establishes interactions between multimodal sequences, causing many invalid tokens (such as irrelevant image regions or redundant text fragments) to participate in computation, resulting in significant memory usage and computational redundancy.

To solve these problems, this work optimizes on two levels:
\begin{itemize}
    \item At the sequence modeling level, a dynamic sequence padding mechanism is introduced, which computes the maximum sequence length (\texttt{max\_length}) for each training batch according to actual input lengths, achieving dynamic alignment of variable-length sequences and avoiding resource waste caused by traditional fixed-length padding (e.g., uniformly padding to 200 tokens).
    \item In the data preprocessing stage, invalid visual tokens are pruned based on prior knowledge. For example, in the LIBERO dataset, right-hand perspective images are verified to have no significant contribution to task execution, so these images are directly removed before input, reducing the number of visual tokens at the source and lowering attention computation complexity and memory consumption.
\end{itemize}

This optimization method significantly improves training efficiency and reduces invalid computation and memory usage without changing the model structure, providing a more scalable training paradigm for large-scale multimodal policy learning.

\begin{figure}[h!]
    \centering
    \includegraphics[width=0.8\linewidth]{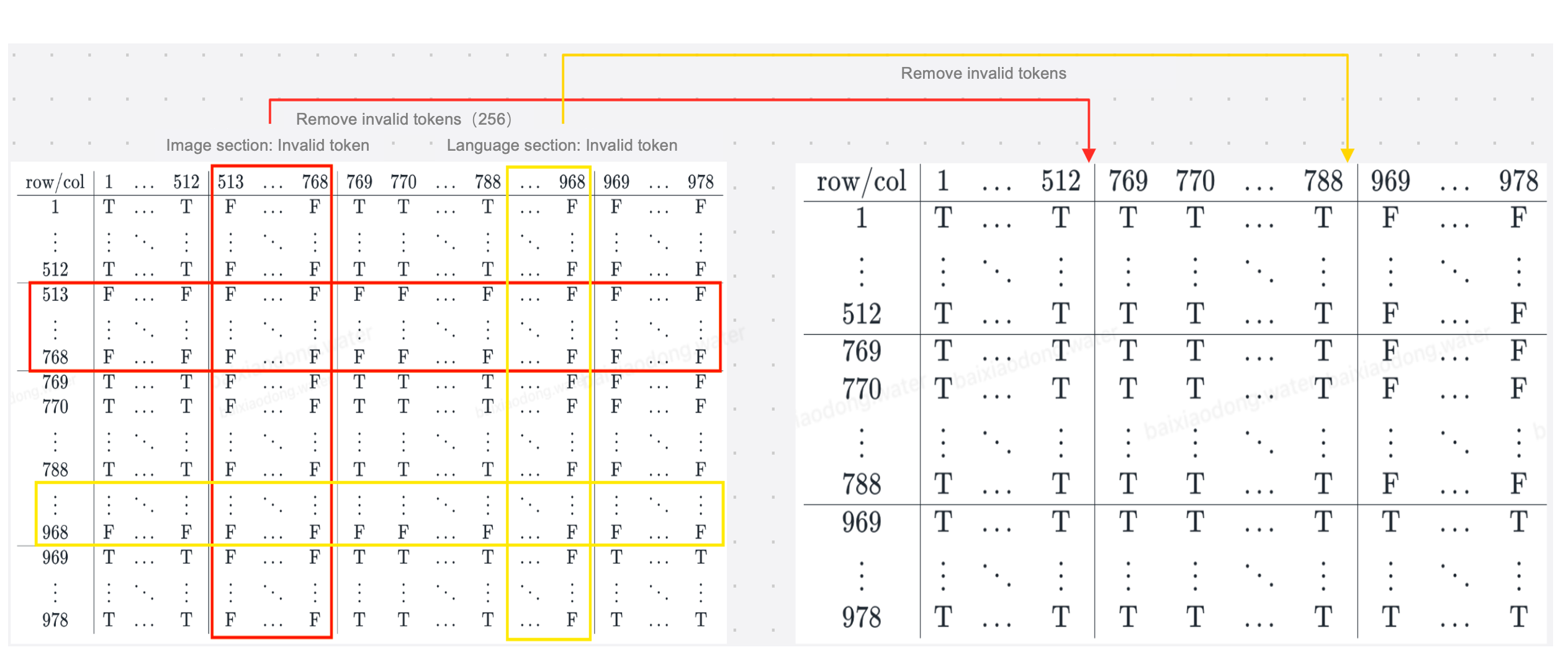}
    \caption{$\pi_{0.5}$ Attention Mask Example}
    \label{fig:placeholder}
\end{figure}

\subsubsection{Quantization Compression: Achieving Model ``Slimming'' and Acceleration While Maintaining Accuracy}

Large model quantization is a model compression technique that reduces model size and accelerates computation through low-bit quantization. The finer the granularity, the smaller the accuracy loss, but managing multiple scaling factors increases computational and memory overhead. Quantization is especially critical for deployment on edge devices and is well-suited for small-parameter VLA models, enabling efficient compression while maintaining accuracy.


The quantization process selectively uses per-tensor, per-channel, or block-wise quantization according to the original neural network layers and data types to maximize model accuracy during conversion.

\begin{itemize}
    \item \textbf{Per-tensor quantization:} Uses a single scaling factor (scalar) to scale the entire tensor.
    \item \textbf{Per-channel quantization:} Each channel of the tensor has a specific scaling factor; for convolutional neural networks, the output channel at the 0th dimension of the kernel is usually the quantization axis.
    \item \textbf{Block-wise quantization:} The tensor is partitioned into fixed-size blocks along one or more dimensions, with a scaling factor defined for each block.
\end{itemize}

The main difference among the three lies in the granularity of scaling. The finer the granularity, the higher the potential accuracy, but managing multiple scaling factors increases computational and memory overhead. Therefore, choosing the appropriate granularity is important to balance the advantages of quantization (e.g., reduced memory usage) and its potential disadvantages (e.g., accuracy loss).

Quantization is crucial for reducing memory usage and improving energy efficiency, making it ideal for deployment on resource-constrained edge devices. This is exactly the scenario faced when deploying VLA/VLM models on terminal devices, making exploration of quantization for VLA/VLM models highly significant.

Ordinary FP8 quantization (i.e., per-tensor scaling) has a significant impact on model accuracy, especially for small models. Due to real-time requirements and edge compute constraints, current VLA and the VLMs they use are usually small models with parameter sizes of 0.5B, 1B, 3B, etc.

We apply fine-grained FP8 quantization (block-wise FP8 quantization), partitioning tensors along the last two dimensions into blocks of size $128 \times 128$. The vision module (ViT) is not quantized to maintain visual feature quality, while the language module (LLM) undergoes fine-grained FP8 quantization, applied post-training (PTQ), without FP8-aware training (QAT). This compresses model size and improves inference speed while maintaining model accuracy.

\begin{figure}[h!]
    \centering
    \includegraphics[width=0.8\linewidth]{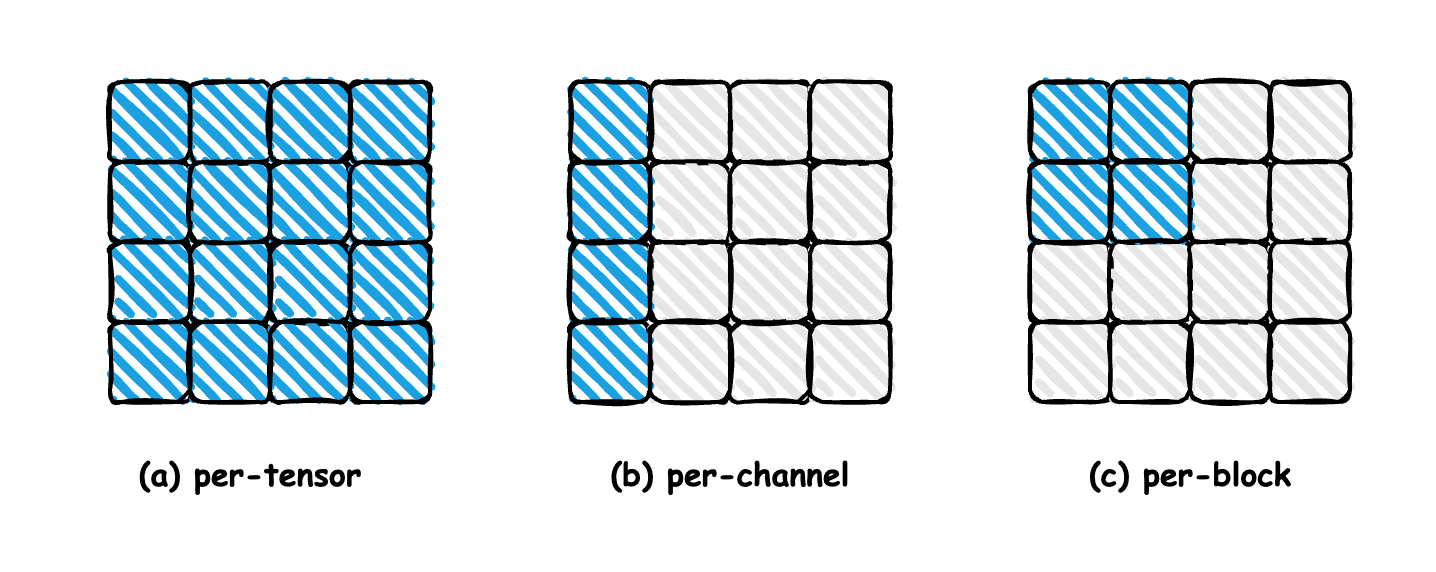}
    \caption{The example of per-tensor, per-channel and per-block}
    \label{fig:placeholder}
\end{figure}

\subsection{RL-VLA$^3$: Reinforcement Learning VLA Accelerating via Full Asynchronism}

Existing VLA training pipelines are inherently constrained by the synchronous execution paradigm, failing to fully leverage the parallel processing capabilities of computational resources. The serial dependencies among simulator interaction, policy generation, and model training lead to idle computational resources and throughput bottlenecks, limiting further improvements in training efficiency. In current practices of LLM reinforcement learning, asynchronous training mechanisms have been proven to significantly enhance training efficiency and system throughput, with related research giving rise to various mature asynchronous optimization strategies. However, in the domain of VLA model reinforcement learning training, such asynchronous training methods remain in the exploratory stage and have not been fully utilized.

\begin{figure}[h!]
    \centering
    \includegraphics[width=0.8\linewidth]{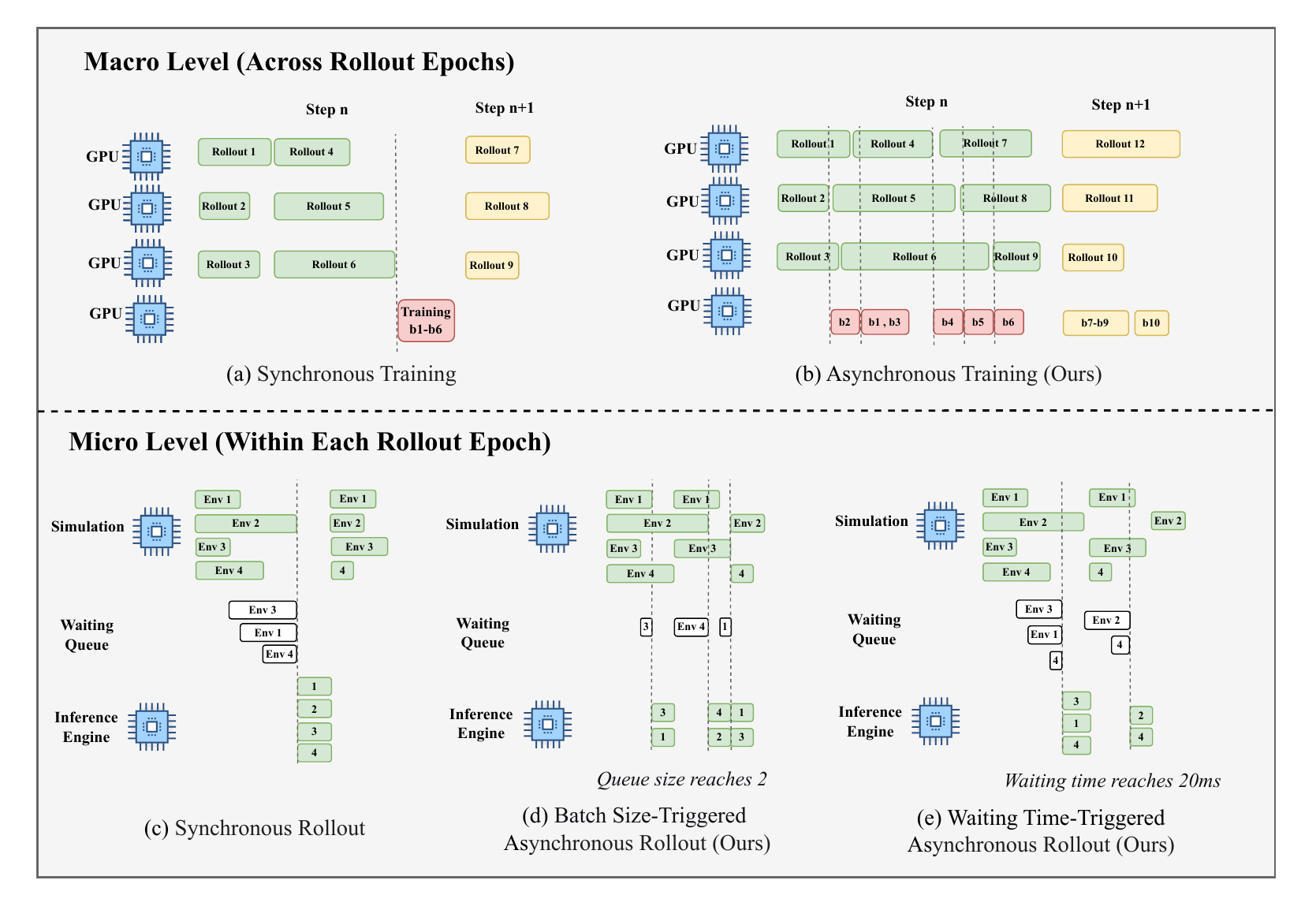}
    \caption{An illustration of the proposed asynchronous framework for VLA training. The design operates asynchronously on two levels: (1) macroscopically, between the rollout and training pipeline stages, and (2) microscopically, within each rollout epoch. The latter is governed by two trigger strategies: batch size and waiting time. }
    \label{rl-vla}
\end{figure}

To bridge these gaps, this paper systematically introduces asynchronous training and inference mechanisms based on an existing unified framework. Drawing on the asynchronous optimization concepts from LLM reinforcement learning, we design a triple-level asynchronous execution architecture, RL-VLA3, which encompasses asynchronous training and inference, asynchronous interaction policy, and streaming generation, as show in Figure~\ref{rl-vla}. These mechanisms substantially reduce inter-component waiting times, enabling continuous and saturated utilization of computational resources. Experiments demonstrate that the proposed asynchronous framework significantly boosts training throughput and reduces overall training time while maintaining good training stability and policy performance.

\subsubsection{Asynchronous training and inference}
Rollout workers (responsible for environment interaction and trajectory generation) and actor workers (responsible for policy model updates) are deployed on entirely different GPU devices. After a rollout worker completes a single trajectory, it immediately places that trajectory into the transmission queue of the communication pipe without waiting for other rollout processes to finish, and proceeds to generate the next trajectory based on the current policy version. The actor worker no longer waits for all rollout processes to complete; instead, once the accumulated trajectory data in the communication pipe reaches a predefined training batch size, it asynchronously collects data from the queue and initiates policy optimization and parameter updates. Asynchronous training and inference achieve computational masking between the two core stages of rollout and actor: the model updates by the actor and trajectory generation by the rollout workers are heavily overlapped in time, thus successfully resolving the resource idle problem of traditional synchronous training.

\subsubsection{Asynchronous interaction policy}
Traditional VLA training frameworks typically employ a synchronous batch interaction method, where multiple environments run in parallel but require all environments to complete their current step before entering model inference as a whole batch, leading to strict synchronization dependencies. RL-VLA$^3$ adopts a dynamic batching scheduler, which balances throughput and latency through two key parameters: 
$B_{max}$ (the maximum batch size allowed for a single inference) and 
$T_{max}$ (the maximum time a request can wait). 
Dynamic batching avoids prolonged idle spinning caused by waiting to form a batch; under high load, it naturally tends toward larger batches to improve throughput, while under low load or environment jitter, it prioritizes system fluidity.

\subsubsection{Streaming generation}
During training, the actor needs to accumulate a sufficient number of trajectory samples to form a complete global training batch before it can start forward and backward computations of the model, leading to intermittent GPU idle periods. RL-VLA3 splits the global training batch into several independent micro-batches. As soon as the accumulated samples reach the size of a single micro-batch, the actor immediately initiates forward and backward computations for that micro-batch. After all micro-batches are computed sequentially, the gradients generated from all micro-batches are aggregated together, and a single model parameter update is performed, thereby avoiding intermittent idle periods.

\section{Experiments}

\subsection{Thousand-GPU Scale Framework Validation}

Based on the aforementioned architecture and cloud-native optimizations, AI Infrastructure has achieved a breakthrough in the scalability and efficiency of embodied intelligence training on JoyBuilder, validating the framework's advancement and reliability.

\subsubsection{DDP Distributed Thousand-GPU Training Scaling Law}

\begin{figure}[htbp]
\centering
\subfloat[Diagram illustrating the relationship between mini-batch size, training time, and GPU memory usage.]{\label{fig:ddp_bs}%
  \includegraphics[width=0.48\linewidth]{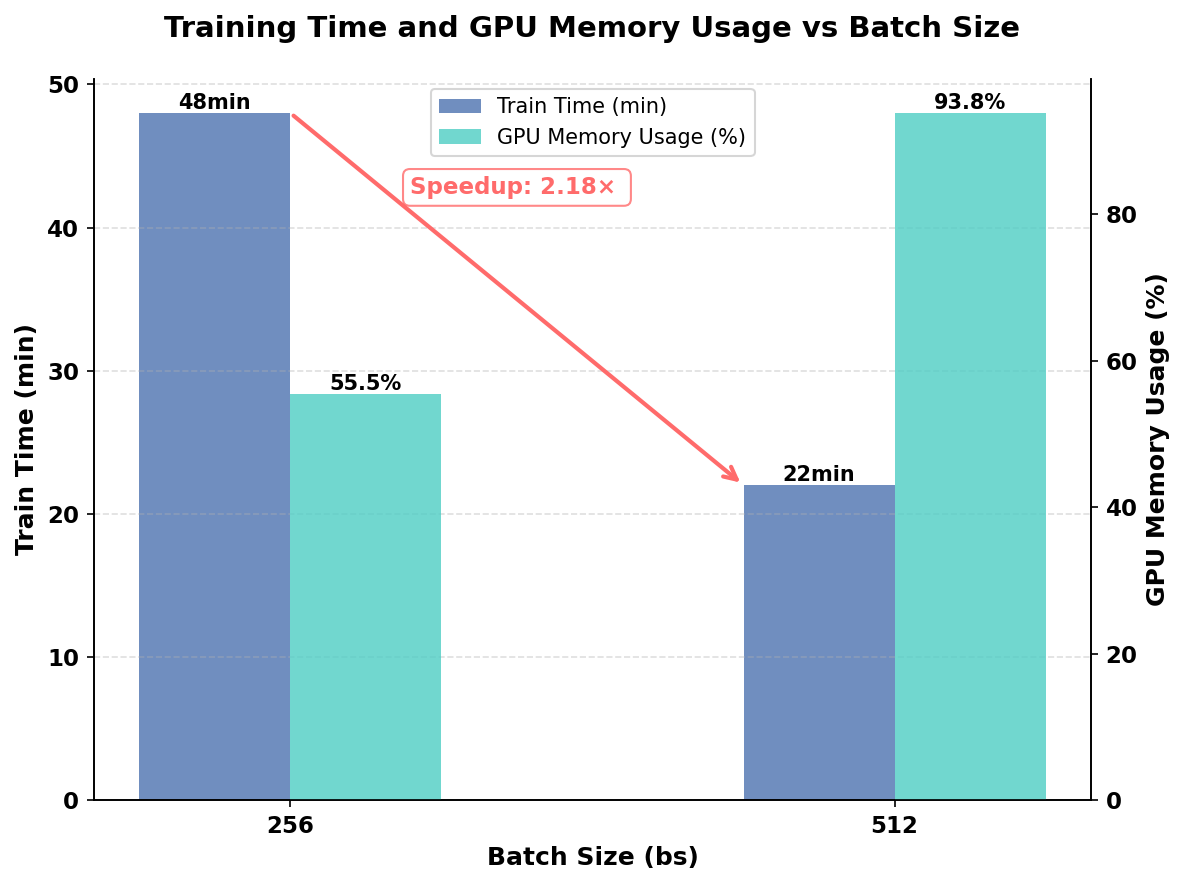}}\hfill
\subfloat[Diagram illustrating the relationship between data parallelism and training time.]{\label{fig:ddp_node}%
  \includegraphics[width=0.48\linewidth]{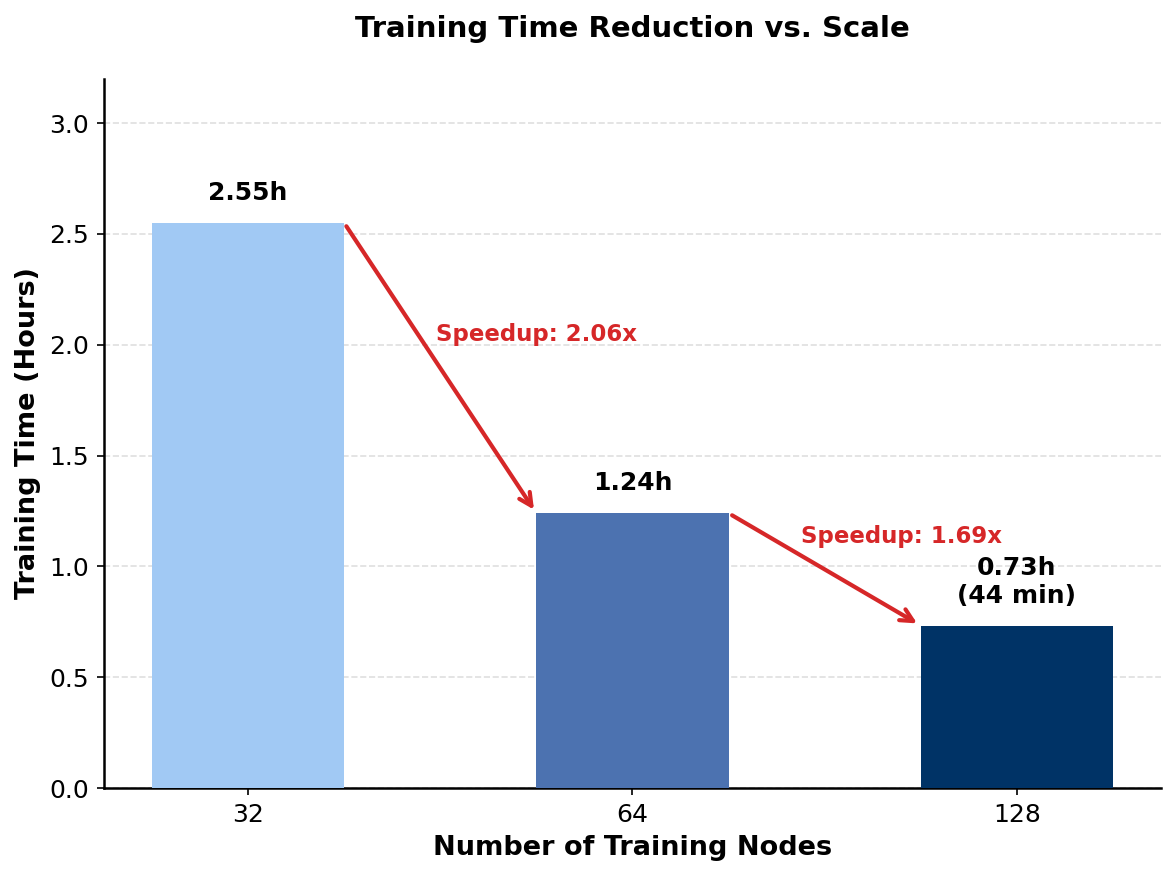}}
\caption{The impact of different mini-batch sizes (MBS) and data parallel instances (DP) on training time and GPU memory usage.}
\label{fig:ddp_scaling}
\end{figure}

An important observation metric for the scaling law in DDP training is the training time under different global batch sizes. Therefore, we focus on exploring the impact of two key factors that constitute the global batch size (GBS): mini-batch size (MBS) and data parallel instances (DP), on training time.

\begin{itemize}
    \item \textbf{Mini-Batch Size (MBS):}
With DP fixed at 128 nodes (1024 GPUs), we compare cases where MBS is 256 and 512. Experimental results show that with MBS=256, the training time per epoch is 48 minutes and memory utilization is 55.5\%. When MBS=512, combined with further storage optimization, the training time per epoch is reduced to 22 minutes and memory utilization rises to 93.98\%, as shown in Figure~\ref{fig:ddp_scaling}. Thus, as MBS increases, training time decreases while memory usage further increases.

\item \textbf{Data Parallel Instances (DP):}
With MBS fixed at 128, we compare training times per epoch for DP at 32 nodes (256 GPUs), 64 nodes (512 GPUs), and 128 nodes (1024 GPUs). Results show that as DP increases, training time per epoch decreases from 2.55 hours to 1.24 hours and 0.73 hours, respectively. Specifically, when DP increases from 32 to 64 nodes, training time is halved, but when DP increases from 64 to 128 nodes, distributed communication overhead also increases, resulting in a final speedup of 1.69x.
\end{itemize}

\subsubsection{Thousand-GPU Training of GR00T N1.5}

\begin{figure}[h!]
    \centering
    \includegraphics[width=0.6\linewidth]{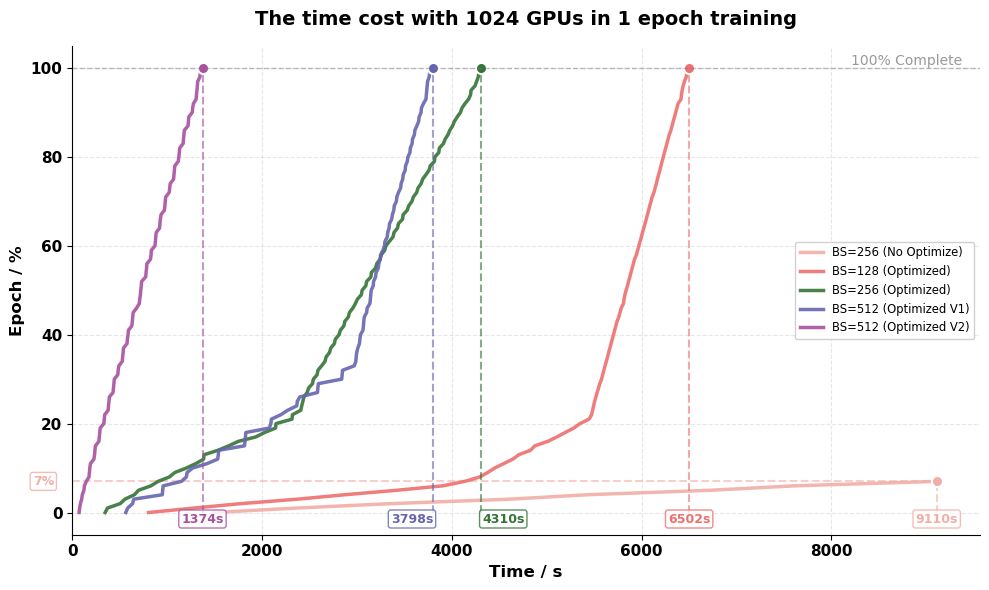}
    \caption{Comparison of system training speeds before and after optimization}
    \label{fig:Speed}
\end{figure}

Leveraging the JoyBuilder embodied toolchain, we achieved stable and efficient training of the GR00T N1.5 model on over 100 million frames of embodied data using a 1024-GPU cluster. Prior to optimization, the maximum supported batch size was 256; exceeding this value would cause heavy file reading, leading to I/O blocking of Dataloader worker processes on certain nodes, which in turn triggered NCCL timeouts and interrupted training. Under these conditions, training for one epoch took approximately 15 hours (as shown by the light pink line in Figure~\ref{fig:Speed}). After end-to-end collaborative optimization of cloud storage, network communication, and platform deployment with JoyBuilder, training with a batch size of 512 required only 22 minutes per epoch (as shown by the purple line in Figure. The training time was reduced by 97.57\%, achieving an approximately 40-fold speedup, truly realizing “fast, accurate, and stable” model training on the platform. Meanwhile, we have established a comprehensive simulation evaluation system, providing precise assessment standards for algorithm iteration.

\begin{figure}[h!]
\centering
\begin{subfigure}[t]{0.48\linewidth}
  \includegraphics[width=\linewidth]{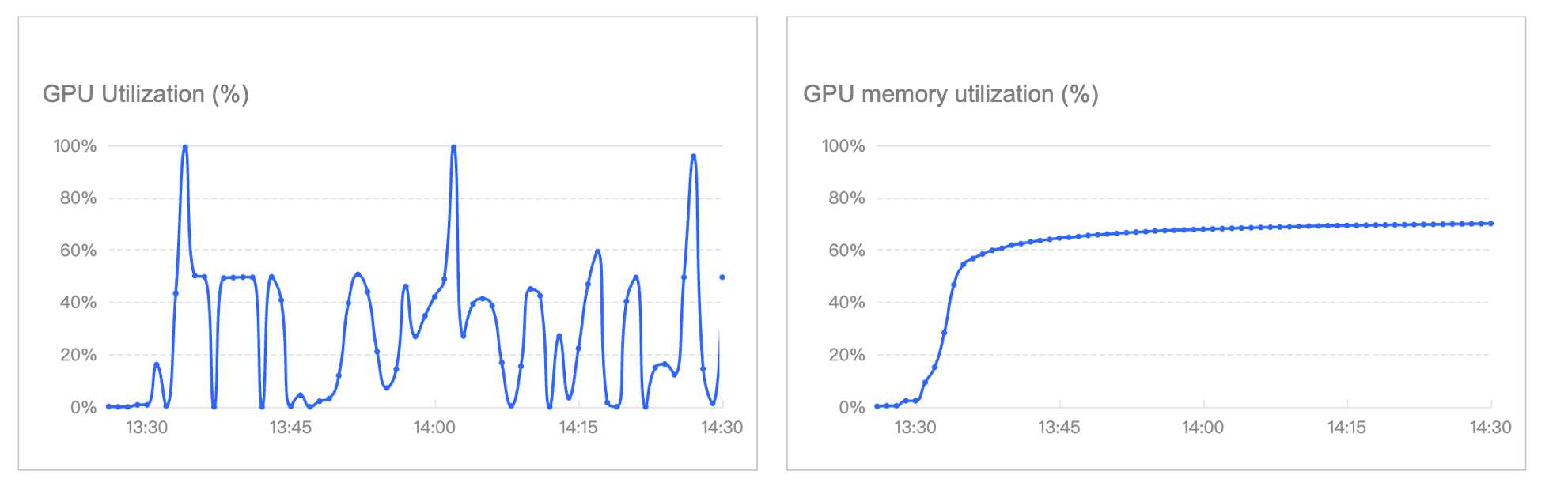}
  \caption{Lerobot Industry-Standard Framework (GR00T N1.5) brfore optimization}\label{fig:joy0}
\end{subfigure}\hfill
\begin{subfigure}[t]{0.48\linewidth}
  \includegraphics[width=\linewidth]{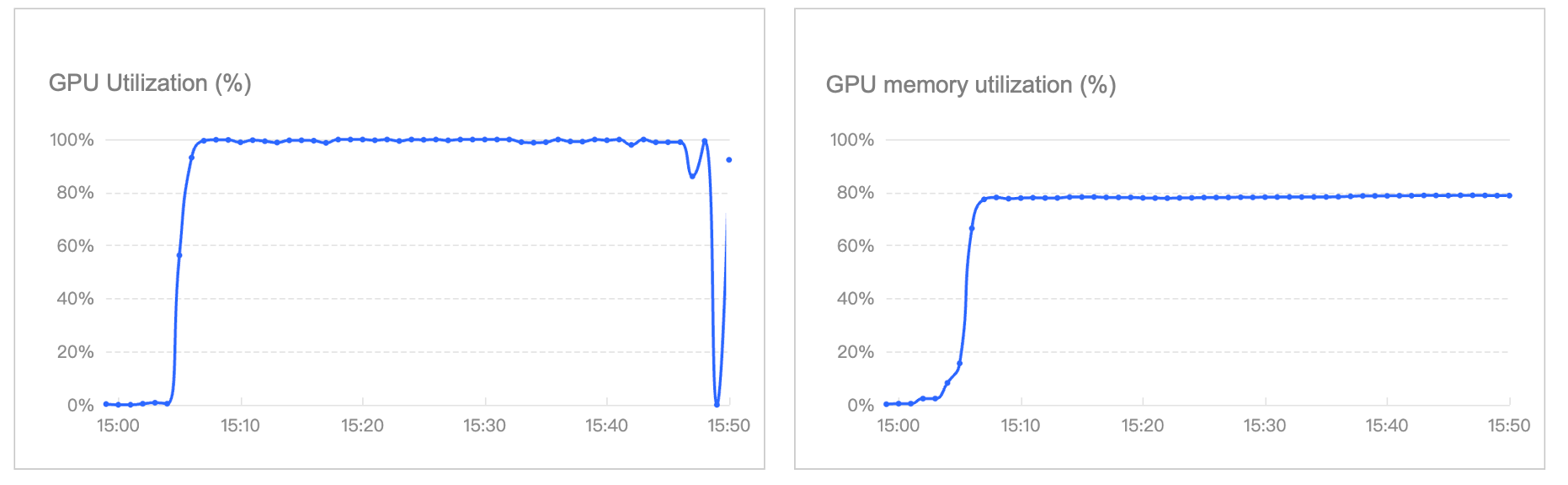}
  \caption{The optimized Lerobot industry-standard framework (GR00T N1.5) has achieved full utilization of both computing power and memory, resulting in high efficiency.}\label{fig:joy1}
\end{subfigure}

\begin{subfigure}[t]{0.48\linewidth}
  \includegraphics[width=\linewidth]{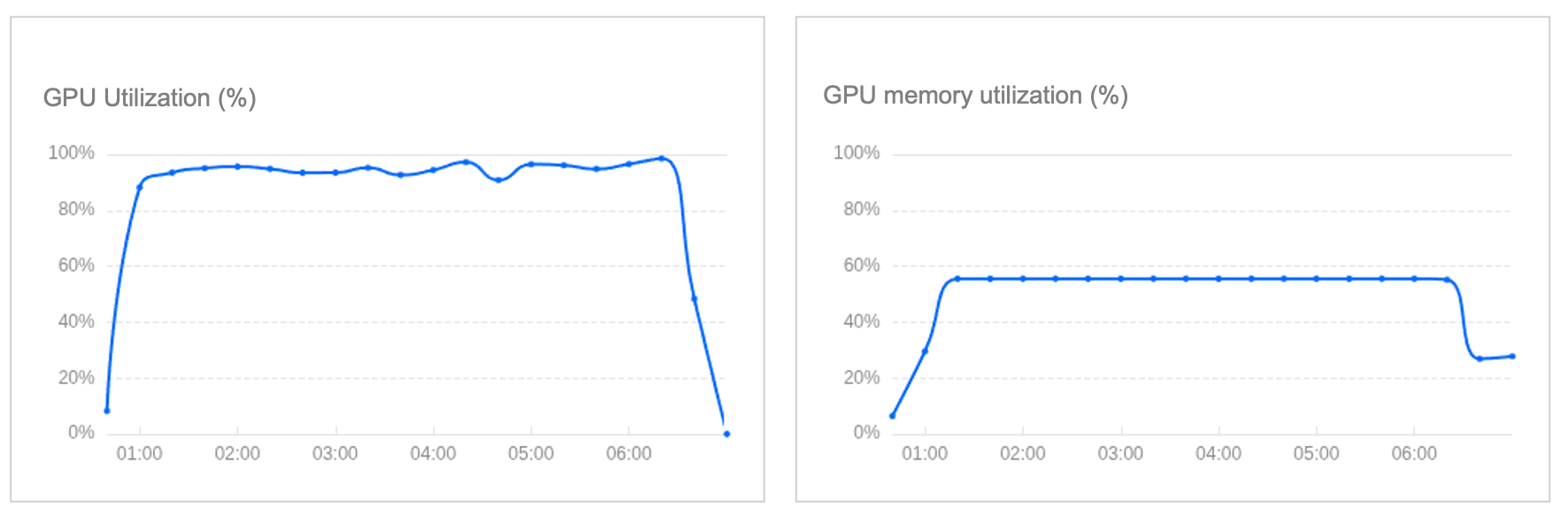}
  \caption{Model optimization pipeline for NVIDIA framework (GR00T N1.5): Before optimization}\label{fig:joy2}
\end{subfigure}\hfill
\begin{subfigure}[t]{0.48\linewidth}
  \includegraphics[width=\linewidth]{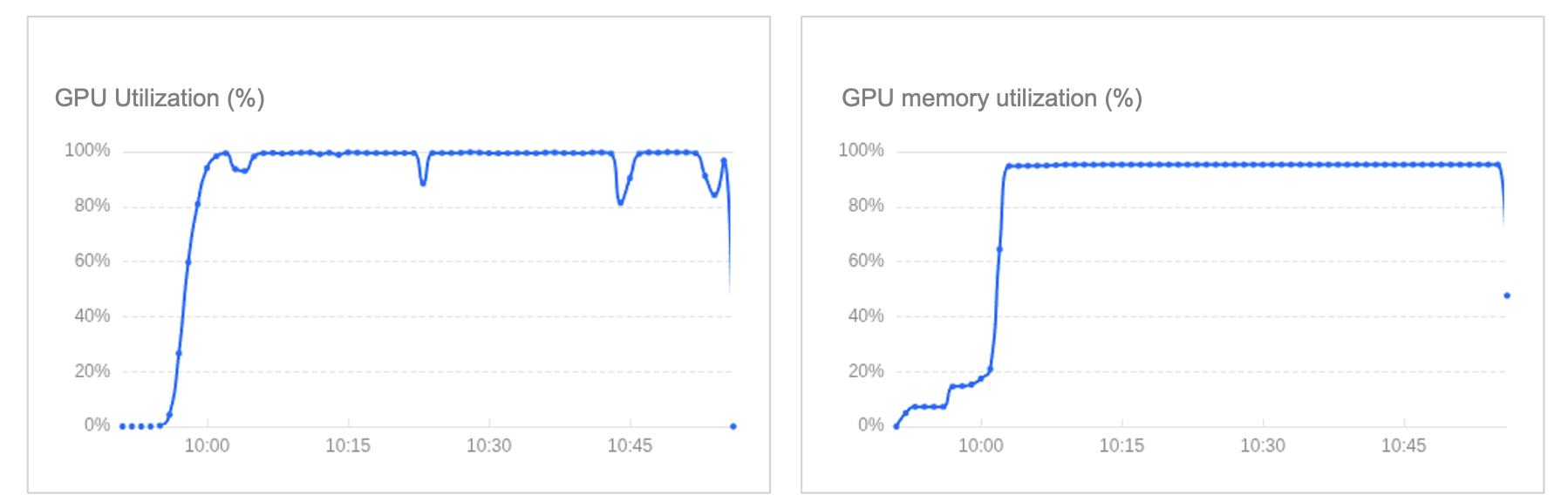}
  \caption{Model optimization pipeline tailored for the NVIDIA framework (GR00T N1.5): improved utilization and stability after optimization.}\label{fig:joy3}
\end{subfigure}

\caption{Comparison of model optimization using the Lerobot industry-standard framework (top) and the NVIDIA-oriented framework (GR00T N1.5) (bottom).}
\label{fig:joy}
\end{figure}

Based on the widely adopted LeRobot open-source framework—which supports the largest variety of embodied models in the industry—and integrating AI Infra’s proprietary cloud-native embodied optimizations, we have achieved the first stable thousand-GPU scale training of the GR00T model in the industry. Training efficiency has been improved by 3.5 times compared to the open-source LeRobot baseline (reduced from 3 hours to 40 minutes, as shown in Figure~\ref{fig:ddp_node}), significantly lowering the barriers and costs for large-scale training and enabling stable thousand-GPU training.

\subsection{Attention Dynamic Computation and Memory Optimization}
\begin{figure}[htbp]
\centering
\begin{subfigure}[t]{0.48\linewidth}
  \includegraphics[width=\linewidth]{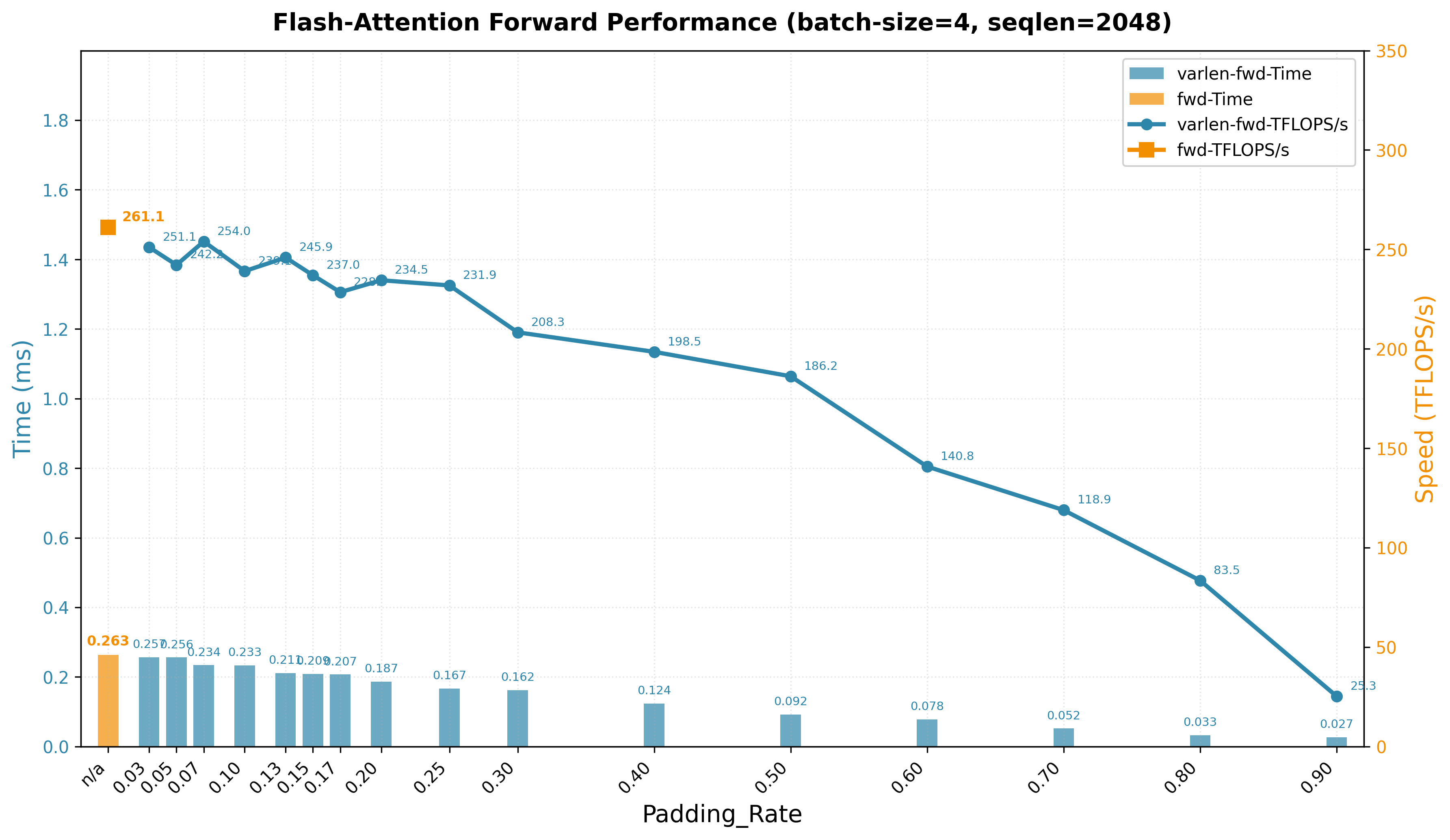}
  \caption{}\label{fig:fig:different_bs_0}
\end{subfigure}\hfill
\begin{subfigure}[t]{0.48\linewidth}
  \includegraphics[width=\linewidth]{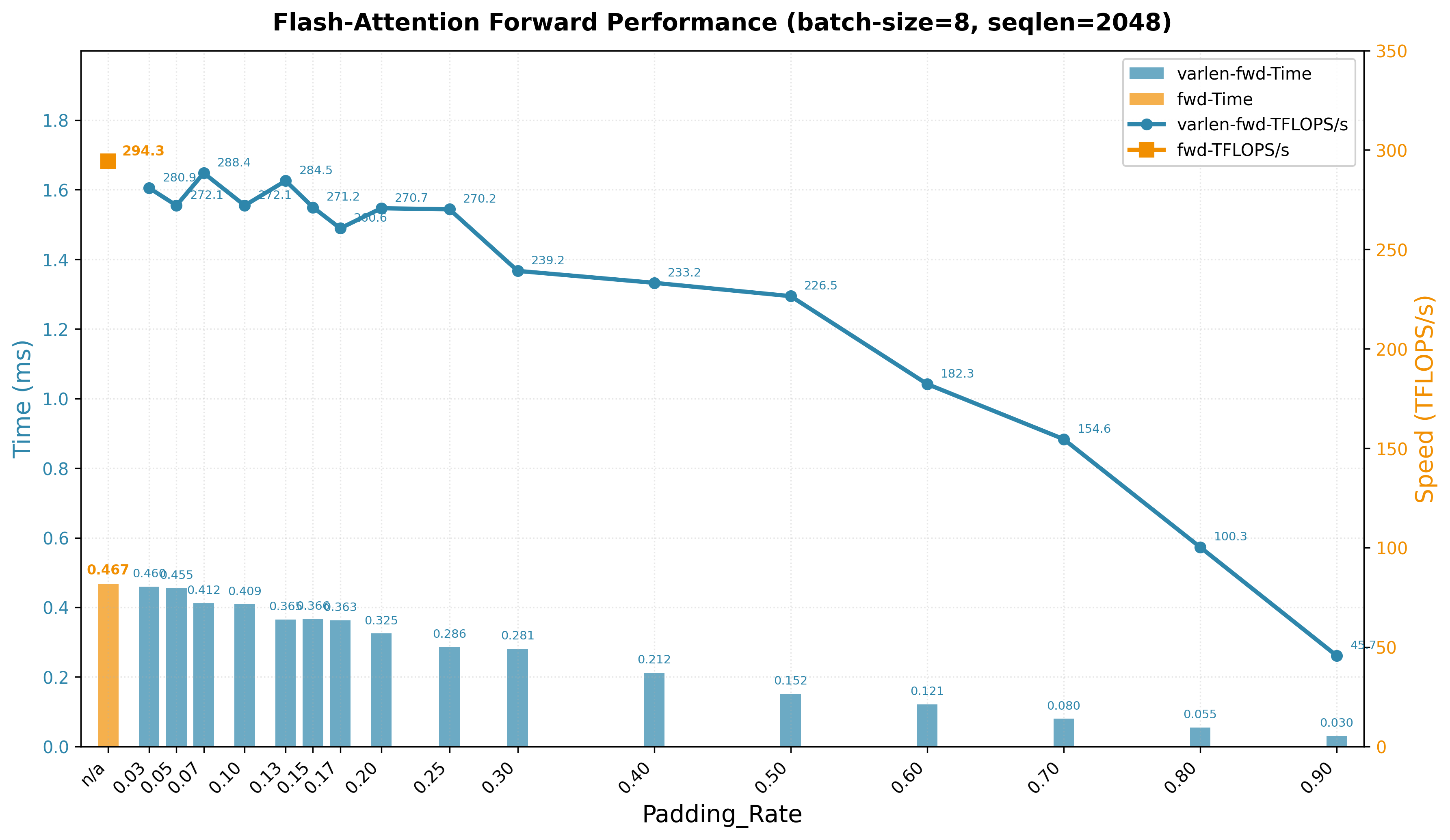}
  \caption{}\label{fig:joy1}
\end{subfigure}

\begin{subfigure}[t]{0.48\linewidth}
  \includegraphics[width=\linewidth]{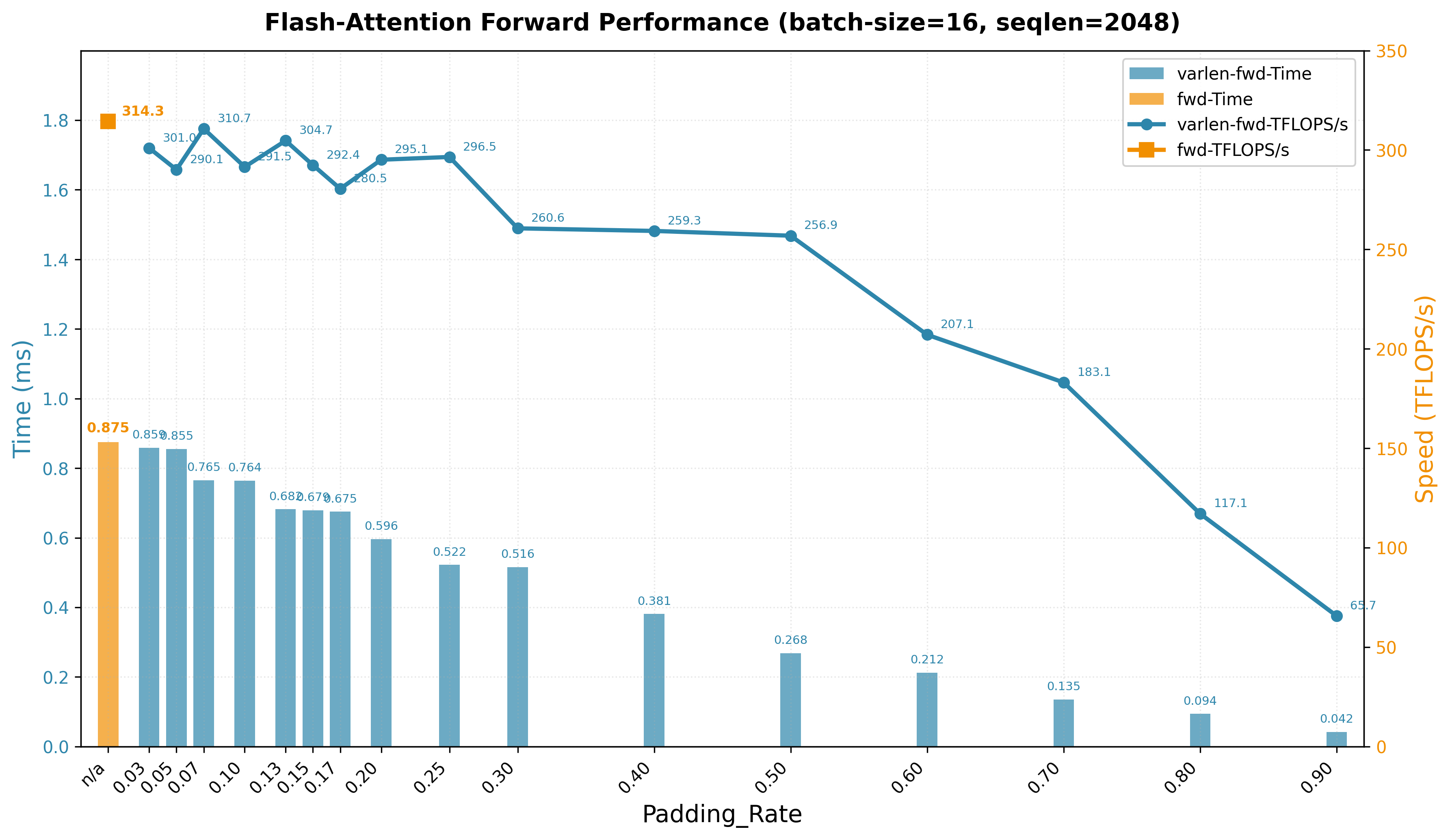}
  \caption{}\label{fig:joy2}
\end{subfigure}\hfill
\begin{subfigure}[t]{0.48\linewidth}
  \includegraphics[width=\linewidth]{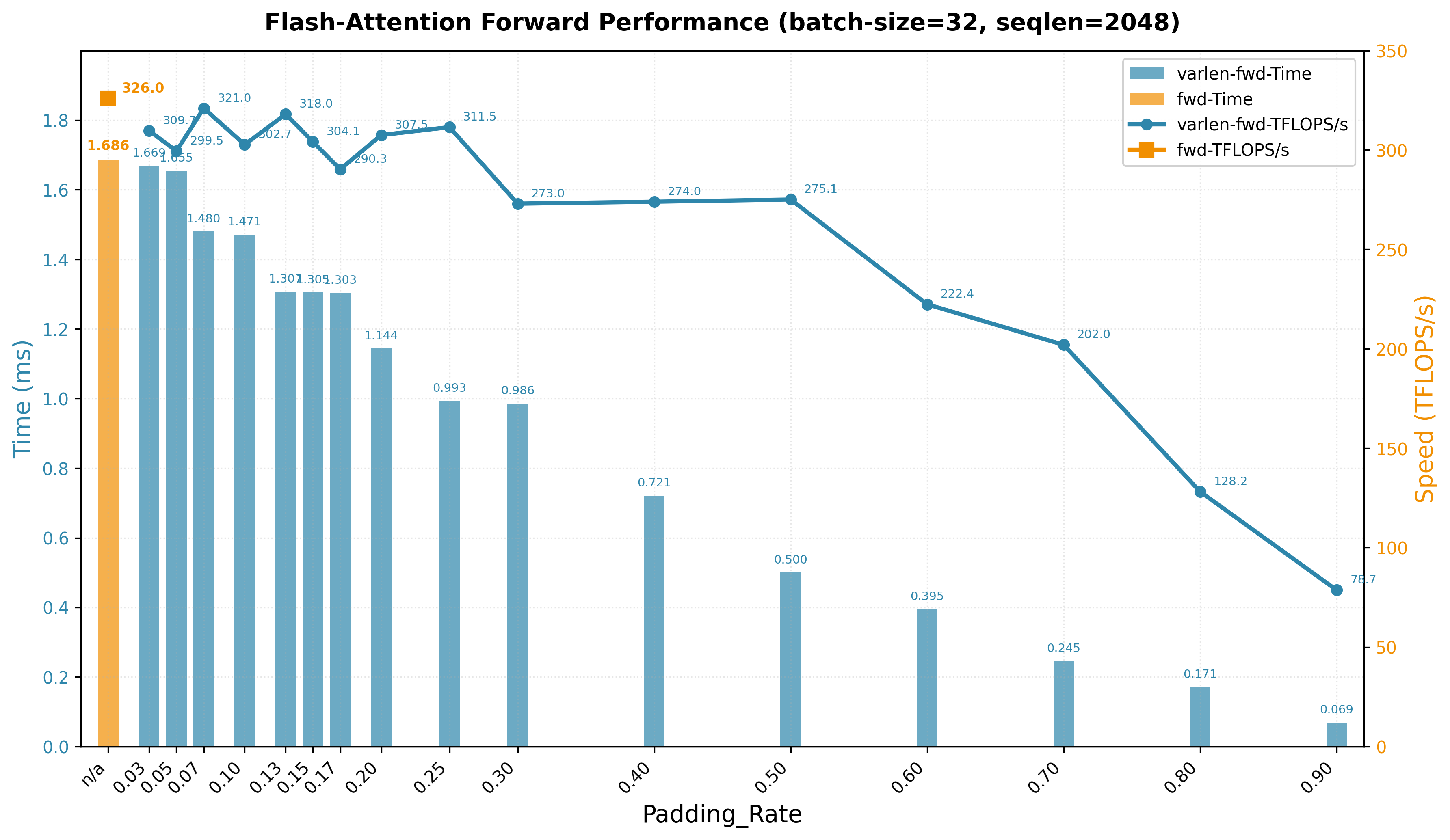}
  \caption{}\label{fig:joy3}
\end{subfigure}

\caption{Different batch size of flash attention.}
\label{fig:different_bs}
\end{figure}
\subsubsection{Variable-Length Flash-Attention: Eliminating Compute Waste from Padding}

\begin{figure}[htbp]
\centering
\subfloat[Comparison of processing time under different sequence lengths and padding rates (batch\_size=16).]{\label{fig:varlen_0}%
  \includegraphics[width=0.33\linewidth]{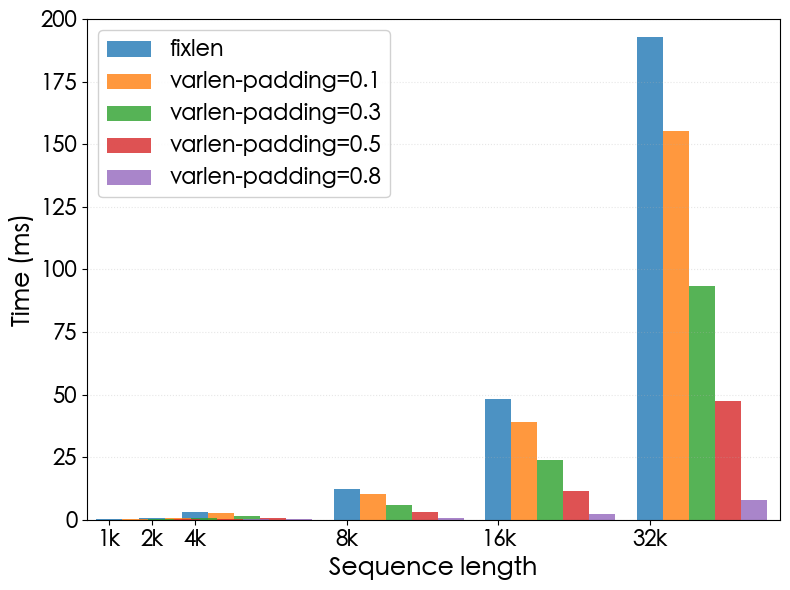}}\hfill
\subfloat[TFLOPS comparison under different sequence lengths and padding rates (batch\_size=16)]{\label{fig:varlen_1}%
  \includegraphics[width=0.35\linewidth]{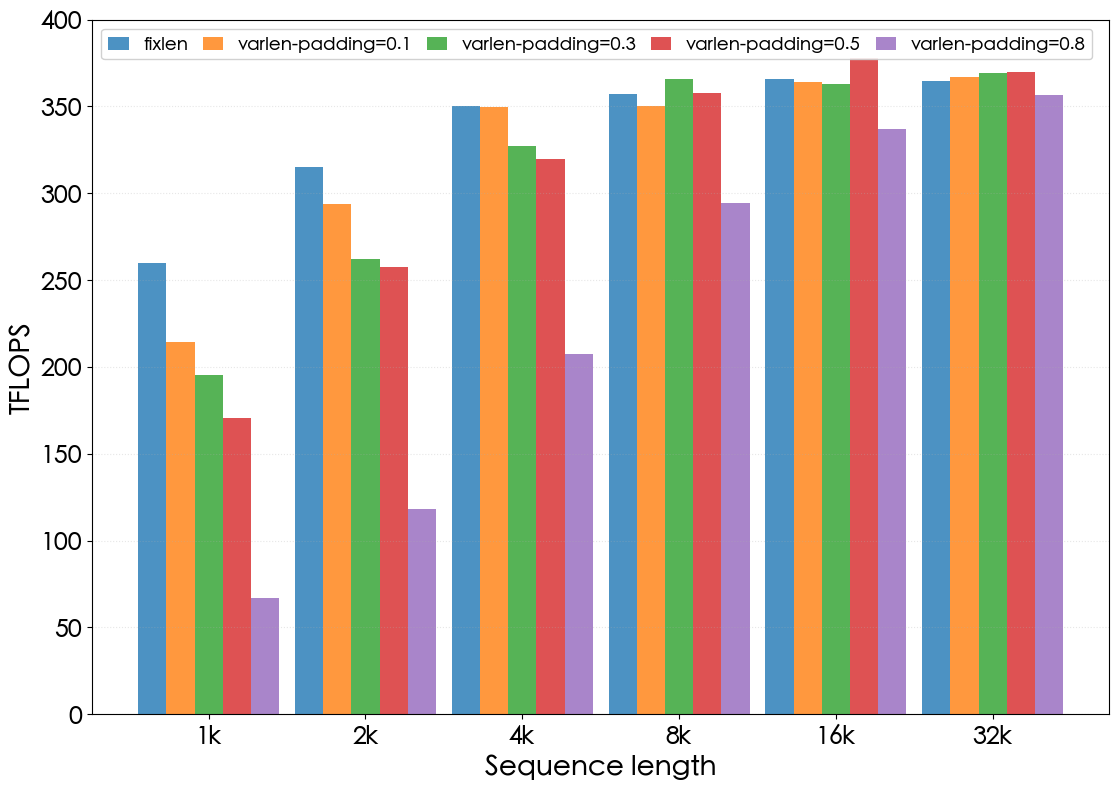}}
\subfloat[Benchmark Score.]{\label{fig:varlen_2}%
  \includegraphics[width=0.25\linewidth]{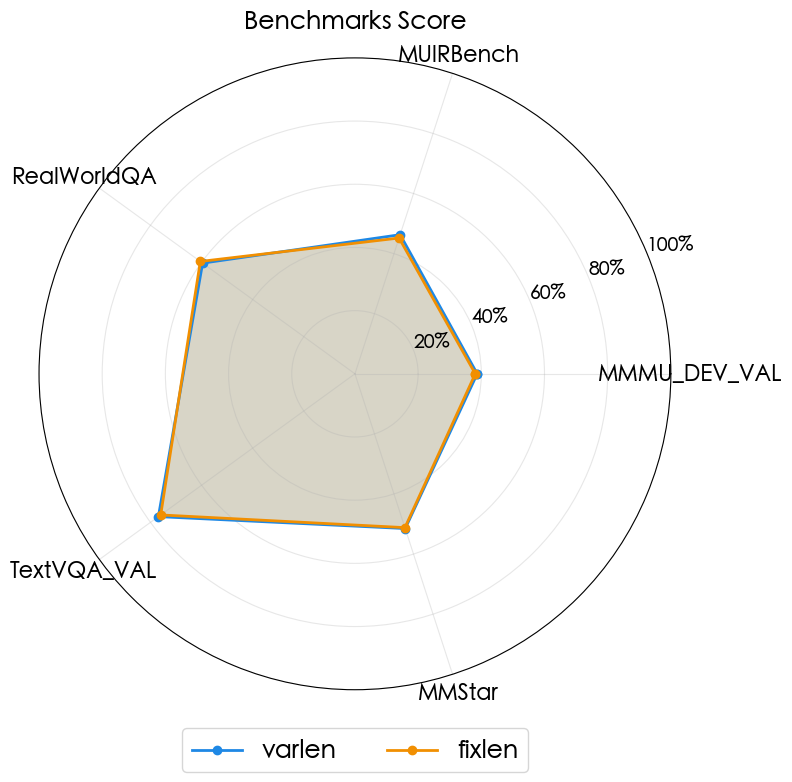}}
\caption{Comparison of efficiency and model performance between variable-length (varlen) and non-variable-length (fixlen) Attention mechanisms}
\label{fig:varlen}
\end{figure}

As the ``understanding core'' of VLA, the VLM undertakes cross-modal understanding tasks for images and text, and its computational efficiency directly determines the training and inference performance of the entire VLA model. In this experiment, Qwen2.5-VL was selected as the base model. Leveraging Flash-Attention with variable-length optimization, computation is performed only on valid token sequences, significantly improving training and inference efficiency while maintaining stable model performance. Experimental results are shown in Figure~\ref{fig:different_bs} and Figure~\ref{fig:varlen}:
\begin{itemize}
    \item When fixing the sequence length (seq\_len) and batch size, examining the impact of different padding rates: In Figure~\ref{fig:different_bs}, for fixed sequence length and batch size, the latency of the variable-length interface (varlen) is consistently lower than that of the fixed-length interface (fixlen). As the padding rate increases from 3\% to 90\%, the time savings from the variable-length interface rise from 2.28\% to 89.73\% (Figure~\ref{fig:fig:different_bs_0}). Furthermore, this advantage becomes more pronounced as the batch size increases.
    For shorter sequences (2048), when the batch size increases from 8 to 32, the peak TFLOPS increases by 1.2x. However, due to the shorter effective sequence lengths, the variable-length interface has lower computational utilization, and its TFLOPS remains lower than the fixed-length interface. In contrast, for long sequences (32k), the variable-length interface executes faster, and its TFLOPS is on par with or even surpasses the fixed-length interface, as shown in Figure ~\ref{fig:varlen_1}.
\item When fixing the batch size (e.g., 16), as the sequence length increases from 2k to 32k, the average latency of the variable-length interface is consistently lower than the traditional method. For sequence lengths larger than 8k, the time savings range from 25\% to 90\%, as shown in Figure~\ref{fig:varlen_0}. The optimization effect of the variable-length interface is particularly significant under higher padding rates.
Additionally, Figure~\ref{fig:varlen_1} shows that TFLOPS increases rapidly as the sequence length grows from 1k to 8k, then stabilizes beyond 8k, indicating that the variable-length interface achieves higher computational efficiency in large-scale sequence processing.
While enabling faster computation, the performance of the variable-length interface remains consistent with the fixed-length interface across multiple benchmarks, as shown in Figure~\ref{fig:varlen_2}.
\end{itemize}

These quantitative results provide measurable engineering optimization evidence for the efficient training of deep learning models.

\subsubsection{Data Packing: From Sample Redundancy to Sequence Integration}
\label{expr:datapacking}

\begin{figure}
    \centering
    \includegraphics[width=0.8\linewidth]{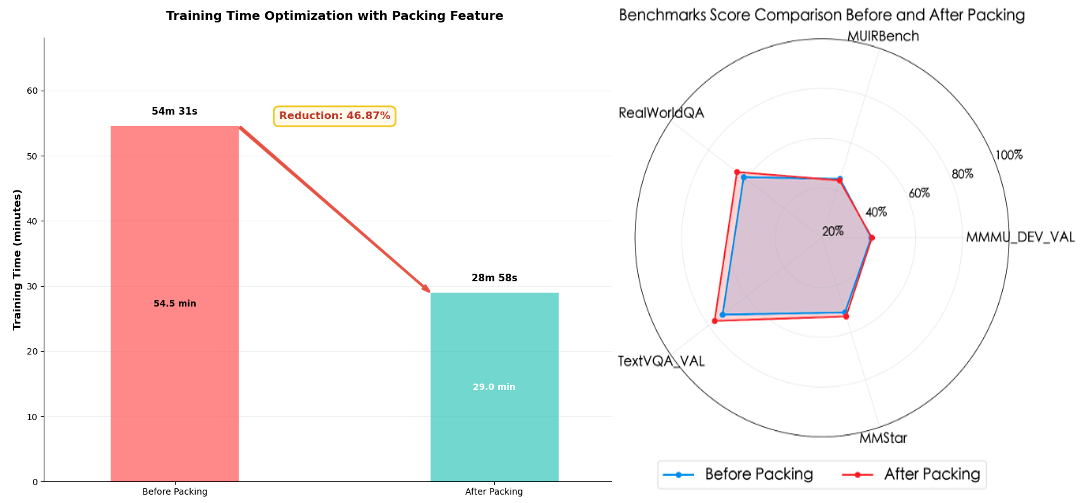}
    \caption{Data Packing combined with the Flash-Attention strategy}
    \label{fig:data_packing_expr}
\end{figure}
To comprehensively evaluate the effect of Data Packing, we assessed its impact on Qwen2.5-VL across multiple datasets, including visual question answering (TextVQA\_VAL), reading comprehension (MMMU\_DEV\_VAL, MUIRBench, RealWorldQA), and multi-domain, multi-language benchmarks (MMStar). Specifically, several shorter samples are concatenated end-to-end to form long sequences close to the model's maximum context length, effectively eliminating invalid padding tokens. Flash-Attention is then used to efficiently perform self-attention computation on these long sequences. Experiments result is shown in figure~\ref{fig:data_packing_expr}. This approach achieves a 1.88$\times$ increase in training throughput and a 46.87\% reduction in total training time, while downstream task accuracy remains the same or slightly improves: on MMMU\_DEV\_VAL, MUIRBench, RealWorldQA, TextVQA\_VAL, and MMStar, accuracy changes from 41.00\% to 41.33\%, 44.88\% to 44.23\%, 61.44\% to 64.97\%, 72.58\% to 76.82\%, and 51.60\% to 53.27\%, respectively. These results indicate that the combination of Data Packing and Flash-Attention not only significantly improves training efficiency and memory utilization but also brings additional performance gains in complex open-domain and visual question answering tasks.
With virtually unchanged accuracy, training speed increased by 188\% and total training time decreased by 46.87\%.

\subsection{$\pi_{0.5}$ Acceleration Optimization}

\begin{figure}[htbp]
\centering
\subfloat[Training results.]{\label{fig:pi_resluts}%
  \includegraphics[width=0.48\linewidth]{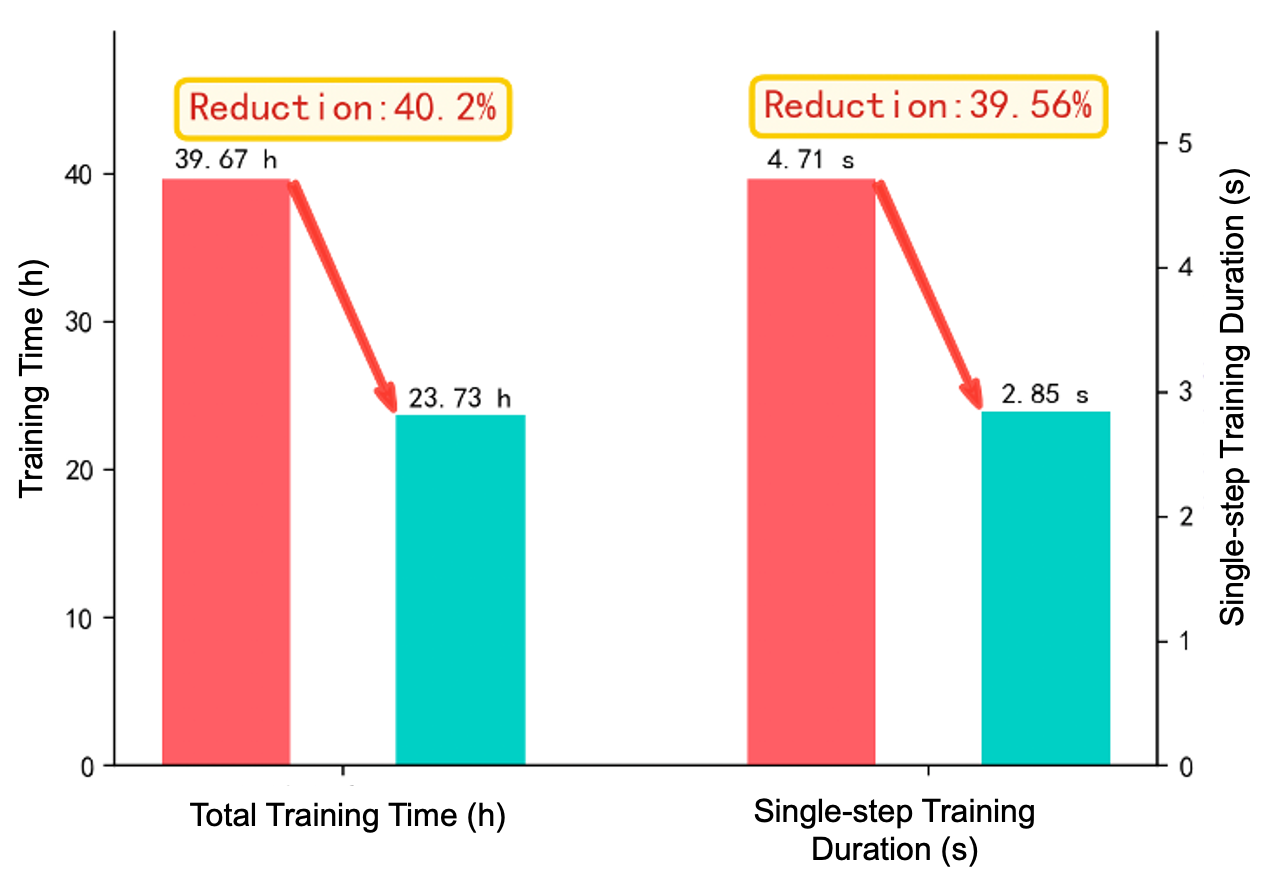}}\hfill
\subfloat[inference result]{\label{fig:pi_inference}%
  \includegraphics[width=0.4\linewidth]{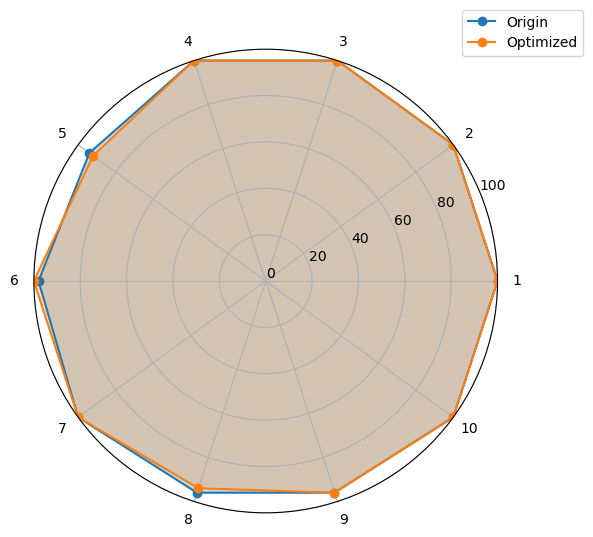}}
\caption{omparison of training duration before and after optimization and verification of simulation results}
\label{fig:pi_expr}
\end{figure}

\begin{table}[htbp]
    \centering
    \caption{Task Success Rate Comparison in $\pi_{0.5}$: Original vs. Optimized}
    \begin{tabular}{cp{7cm}cc}
        \hline
        No. & Task Name & Original & Optimized \\
        \hline
        1 & pick up the black bowl between the plate and the ramekin and place it on the plate & 100\% & 100\% \\
        2 & pick up the black bowl next to the ramekin and place it on the plate & 100\% & 100\% \\
        3 & pick up the black bowl from table center and place it on the plate & 100\% & 100\% \\
        4 & pick up the black bowl on the cookie box and place it on the plate & 100\% & 100\% \\
        5 & pick up the black bowl in the top drawer of the wooden cabinet and place it on the plate & 94\% (3 failures) & 92\% (4 failures) \\
        6 & pick up the black bowl on the ramekin and place it on the plate & 98\% (1 failure) & 100\% \\
        7 & pick up the black bowl next to the cookie box and place it on the plate & 100\% & 100\% \\
        8 & pick up the black bowl on the stove and place it on the plate & 96\% (2 failures) & 94\% (3 failures) \\
        9 & pick up the black bowl next to the plate and place it on the plate & 96\% (2 failures) & 96\% (2 failures) \\
        10 & pick up the black bowl on the wooden cabinet and place it on the plate & 100\% & 100\% \\
        \hline

    &Total success rate  & 0.984  & 0.982 \\
    &Total episodes & 500 & 500\\
    \hline
    \end{tabular}
    \vspace{0.5em}
    \label{table:pi_infer}
\end{table}

In this study, we systematically optimized model training efficiency and compared performance before and after optimization through 30{,}000-step fine-tuning experiments on the Libero dataset. Experimental results in Figure~\ref{fig:pi_resluts} show that, during training, the optimized model's per-step training time was reduced from 4.71 seconds to 2.85 seconds, a decrease of 39.56\%. The total training time shortened from 39 hours 40 minutes to 23 hours 44 minutes, representing a 40.2\% improvement in efficiency. Notably, this significant acceleration was achieved while maintaining stable model accuracy: the loss value increased only slightly from 0.0058 to 0.0060, a difference of less than 0.02\%, and deployment success rate remained unchanged. These results demonstrate that the customized optimization strategy achieved a breakthrough in training efficiency without compromising model performance.

To verify the generalization ability of the optimized model, we conducted systematic inference experiments on the Libero Spatial test set. The test set includes 10 independent tasks, each executed 50 times (for a total of 500 rollouts). Results in Figure~\ref{fig:pi_inference} and Table~\ref{table:pi_infer} show that the pre-optimization model succeeded in 492 out of 500 rollouts (8 failures, 98.4\% success rate), while the optimized model succeeded in 491 out of 500 rollouts (9 failures, 98.2\%), with only a marginal difference of 0.2\%. Notably, in Task 5, which had the most failures, the final actions still completed the intended objectives correctly despite some intermediate deviations. Taken together with the training phase improvements (40\% faster training and only a 0.0002 increase in loss) and robust inference performance, these findings confirm that the proposed optimization method significantly improves training efficiency while maintaining model performance ($p > 0.05$, statistically insignificant). This provides strong empirical evidence for efficient deep learning model optimization, showing that increased computational efficiency does not necessarily come at the cost of model accuracy.

\subsection{Post-Training Quantization}

\begin{figure}[h!]
  \centering
  \includegraphics[width=0.48\linewidth]{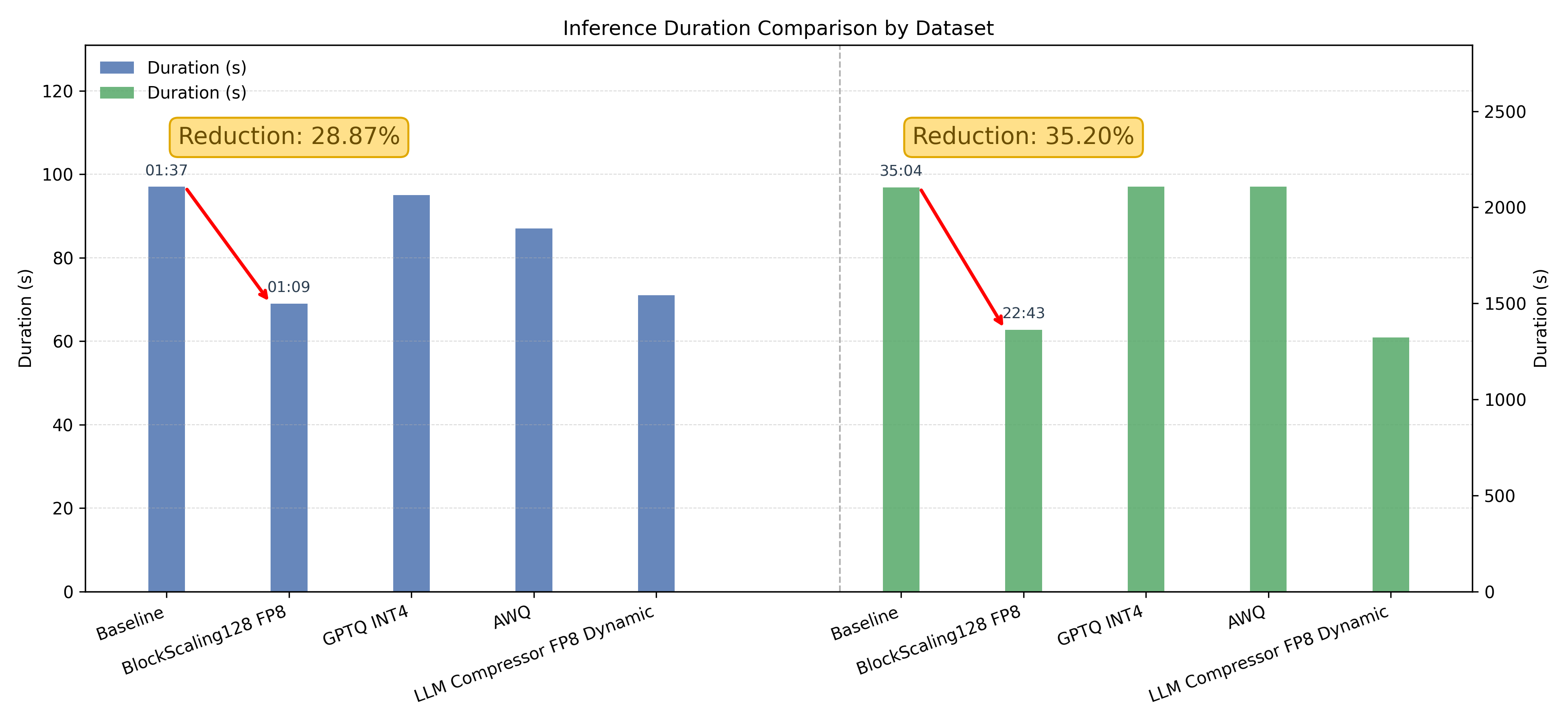}\hfill
  \includegraphics[width=0.48\linewidth]{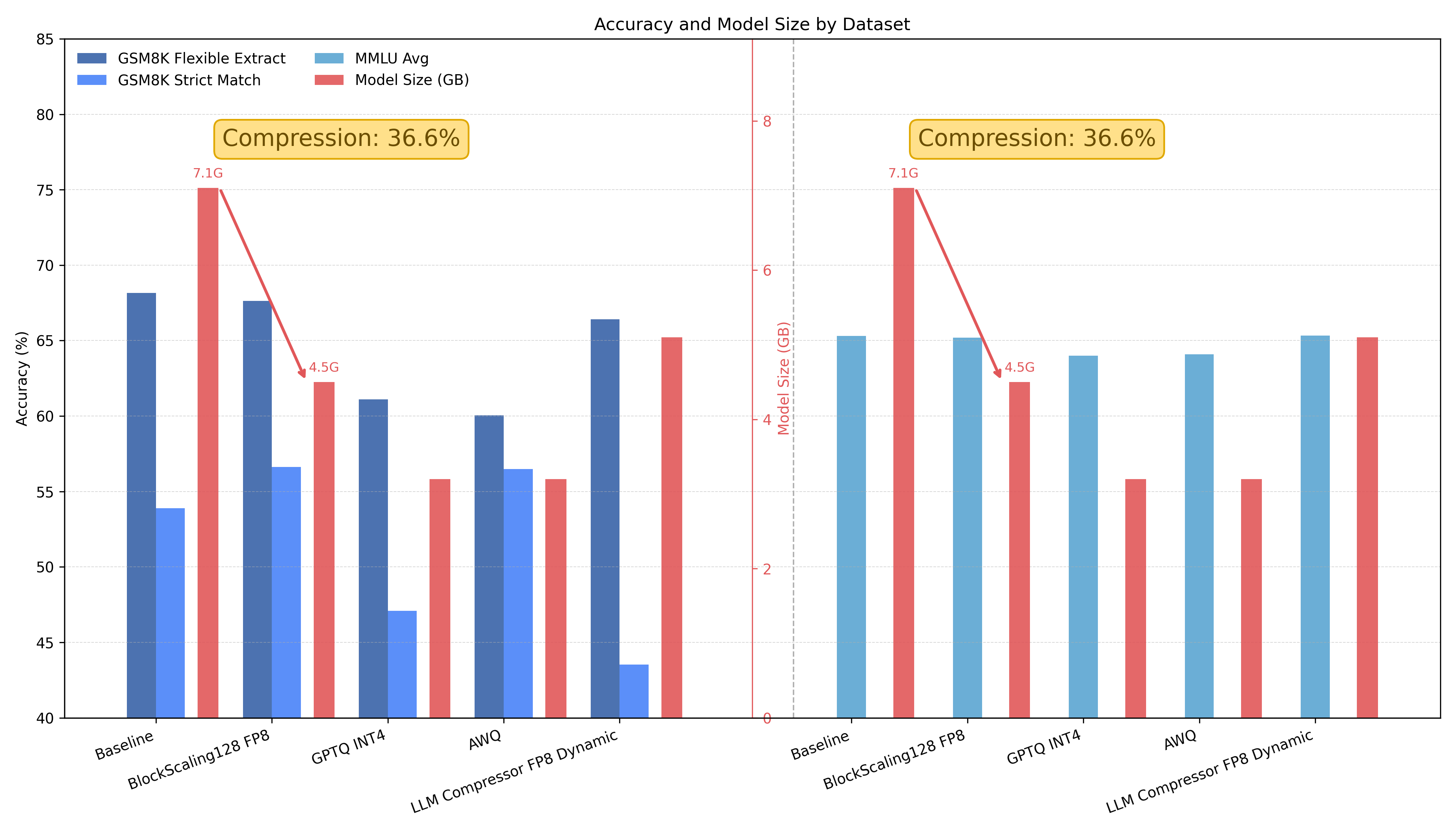}
  \caption{Performance of various quantization models on GSM8K and MMLU tasks}
  \label{fig:quanti_reslut}
\end{figure}

Taking Qwen2.5-VL-3B as an example, the vision module (ViT) retains high precision, while the language module (LLM) adopts fine-grained FP8 block-wise quantization (128$\times$128 blocks), achieving post-training quantization (PTQ) and effectively improving computational efficiency.

We compared various quantization strategies on GSM8K and MMLU tasks, including AWQ, GPTQ-int4, LLM Compressor FP8-dynamic, and our customized FP8 block-wise quantization, as show in Figure~\ref{fig:quanti_reslut}. We can see that FP8 block-wise quantization not only achieved 36.6\% model compression, but also maintained the original model accuracy on both tasks, with post-training quantization speeding up computation by more than 140\%. Overall, FP8 block-wise quantization outperformed other methods in both speed and leaderboard accuracy, achieving ``faster and better'' results.

\subsection{Reinforcement Learning VLA Accelerating via Full Asynchronism}
To demonstrate the performance of RL-VLA3, we conduct experiments measuring the throughput of $\pi_{0.5}$, $\pi_{0}$, and GR00T in the LIBERO and ManiSkill environments. Baseline comparisons are made against the RLinf framework using both co-located and distributed placement strategies.

\begin{table}[h!]
\centering
\renewcommand{\arraystretch}{1.3} 
\caption{Throughput comparison of training strategies. Our asynchronous methods (Train Async, Rollout Async) outperform the Disaggregated baseline and, in most cases, the Colocated baseline across different models and environments. The best value for each column is \textbf{bolded}. And the Increase\% is compare to colocated stratedies. }
\label{tab:performance_comparison}
\resizebox{\textwidth}{!}{
\begin{tabular}{lccccccccc}
\toprule
\multirow{2}{*}{Configuration} & \multicolumn{3}{c}{LIBERO+$\pi_{0.5}$} & \multicolumn{3}{c}{LIBERO+GR00T N1.5} & \multicolumn{3}{c}{ManiSkill+$\pi_{0}$} \\
\cmidrule(lr){2-4} \cmidrule(lr){5-7} \cmidrule(l){8-10}
                               & 8 GPUs & 16 GPUs & 32 GPUs & 8 GPUs & 16 GPUs & 32 GPUs & 8 GPUs & 16 GPUs & 32 GPUs \\
\midrule
Colocated     & 289.23 & 547.55  & 703.85  & 371.80 & 680.46  & 1125.62 & \textbf{132.56} & 232.23  & 370.26  \\
Disaggregated (1:1)       & 162.75 & 307.84  & 457.23  & 220.81 & 409.60  & 729.98  & 60.64  & 150.59  & 257.21  \\
\midrule
\addlinespace[0.4em]
\rowcolor{lightgray}\multicolumn{10}{l}{\textit{Asynchronous improvements on Disaggregated (cumulative):}} \\
\addlinespace[0.4em]
+ Train Async         & 229.68 & 441.49  & 737.46  & 243.57 & 477.20  & 951.33  & 126.65 & \textbf{244.61}  & \textbf{436.32}  \\
+ Rollout Async & 369.56 & 686.80  & 1041.36 & 434.20 & 816.48  & \textbf{1620.39} & 69.07  & 139.32  & 275.36  \\
+ Streamer               & \textbf{383.40 }  & \textbf{713.38}  & \textbf{1120.91} & \textbf{439.64} & \textbf{816.48}  & 1592.40 & 71.61  & 141.49  & 280.07  \\
\midrule
\textbf{Increase \%} & \textbf{$\uparrow$32.5\%} &\textbf{$\uparrow$30.29\%} &\textbf{$\uparrow$59.25\%} & \textbf{$\uparrow$ 18.25\%} & \textbf{$\uparrow$19.99\%} & \textbf{$\uparrow$ 43.96\%} & \textbf{$\downarrow$-4.46\%} &\textbf{$\uparrow$ 5.33\%} & \textbf{$\uparrow$17.84\%}
\\
\bottomrule
\end{tabular}
}
\label{tab:throughput}
\end{table}

As shown in Table~\ref{tab:throughput}, our asynchronous strategy significantly improves throughput compared to the distributed strategy and surpasses the co-located strategy in most scenarios. Training asynchronously (Train Async) delivers substantial throughput gains across all configurations. However, the effectiveness of the asynchronous interaction strategy (Rollout Async) varies across environments. In the LIBERO+$\pi_{0.5}$ environment, this strategy further boosts training efficiency by approximately 40\%. Conversely, it leads to significant performance degradation on ManiSkill. This discrepancy stems from ManiSkill's ability to leverage GPU parallelization for environment computations. Our current Rollout Async implementation partitions the entire batch of environments into mini-batches to achieve genuine parallelism between Environment and Rollout workers, which diminishes the efficiency of batched environment computation in ManiSkill. Nevertheless, the throughput loss is only 4.46\% on a small 8-GPU setup. As the scale increases, the environmental overhead introduced by ManiSkill is offset. At the 32-GPU scale, our method achieves a 17.84\% throughput improvement, validating its effectiveness for large-scale deployment.

\begin{figure}[h!]
    \includegraphics[width=0.8\linewidth]{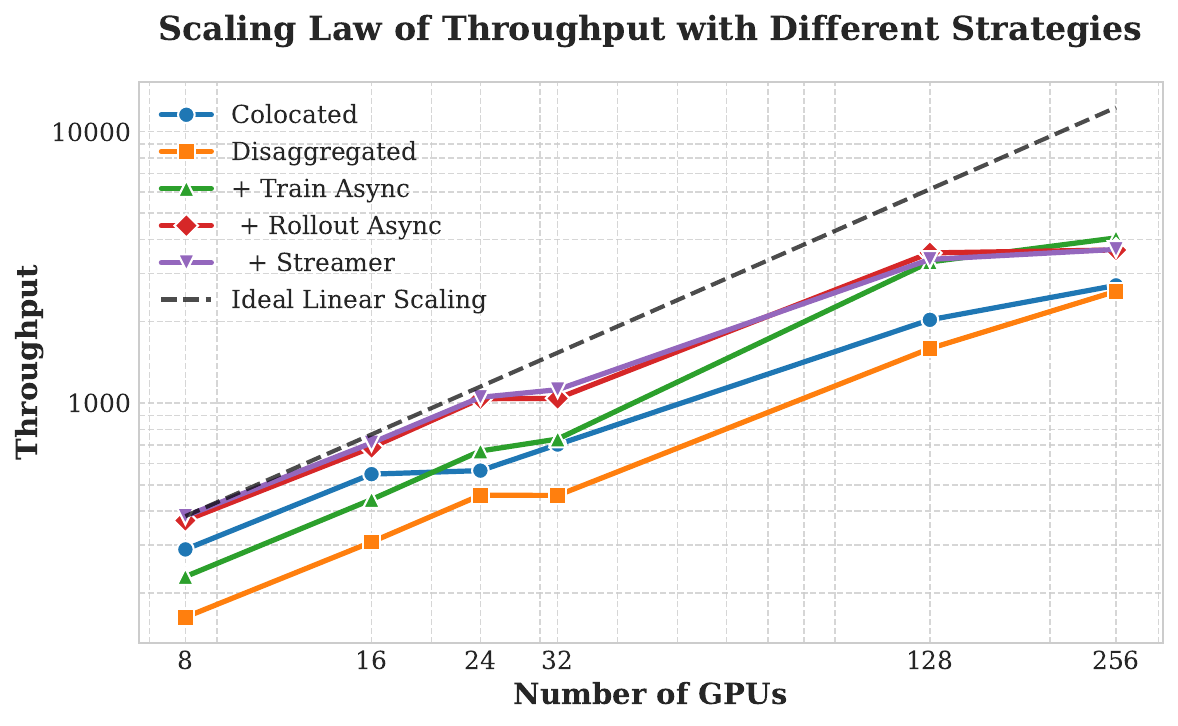}
    \caption{Scaling behavior with increasing GPU resources for the LIBERO+$\pi_{0.5}$}
    \label{fig:scaling_law}
\end{figure}

We further investigate the scaling of different strategies by increasing the number of GPUs. Figure~\ref{fig:scaling_law} shows the throughput for the LIBERO+$\pi_{0.5}$ configuration across different GPU counts. Ideally, throughput should scale linearly with the number of GPUs. Our results indicate that the method exhibits near-optimal scaling performance within the range of 8 to 24 GPUs. Scaling efficiency decreases when scaling from 24 to 128 GPUs and degrades further between 128 and 256 GPUs. The sublinear scaling observed at large scales is attributed to increased communication overhead with the growing number of workers. Improving the method to achieve more desirable scaling capabilities at extreme scales remains an important direction for future work.

\section{Conclusion and Future Outlook}

We have systematically built a thousand-GPU distributed training framework for embodied intelligence targeting VLA models, achieving full-stack breakthroughs from data pipelines to infrastructure. The constructed thousand-GPU training framework has realized systematic optimization at the data, model, and infrastructure layers, and has successfully supported large-scale training for multiple models, including GR00T N1.5. We plan to further extend this to broader model families such as $\pi_0$, continuously validating the framework's advancement and generality.

In the next phase of our research, we will tackle four core challenges inherent to the VLA framework:
\begin{itemize}
    \item Balancing model compactness and expressive power to address inference latency and resource constraints;
    \item Establishing a complete end-to-end Reinforcement Learning infrastructure, spanning from pre-training to reinforcement learning, and constructing a system that integrates world models to achieve a synergistic combination of training, inference, and simulation;
    \item Breaking through the Sim2Real transfer bottleneck by establishing a consistency verification mechanism between simulation and real-world environments;
    \item Improving multimodal evaluation standards to promote the industrialization of embodied intelligence beyond the laboratory.
\end{itemize}

Embodied intelligence is a key area for the deep integration of AI and the physical world. We believe that through continuous system optimization and algorithmic innovation, general embodied intelligence will gradually mature, achieving breakthroughs in autonomous execution of complex tasks by robots, and ultimately ushering in a new stage of human-machine collaboration. One of our future research themes is exploring how to leverage reinforcement learning \cite{junwu2022digitalhumaninteractiverecommendation}, multi-agent systems \cite{wu2025socialitbenchmarkllms}, and the reasoning capabilities of large models \cite{lingteam2025ringlitescalablereasoningc3postabilized}—combined with data synthesis \cite{wu2025sharpsynthesizinghighqualityaligned} and economic design \cite{ma2025learning}—to augment sample data for embodied AI. In addition, whether certain security vulnerabilities inherent in LLMs could be further exacerbated in embodied intelligence is also a concern worthy of our attention~\cite{deng2025ai}. Furthermore, we aim to ensure alignment with human values \cite{jiang2024hummer} and mitigate the potential risks posed by embodied AI to humans \cite{cui2024risktaxonomymitigationassessment}.

\newpage

\section*{Contributions}




\textbf{Author list$^{*}$}\\

Chenfeng Gu$^{1,5}$,
Chen Zhou$^{1}$,
Haoran Li$^{1,3}$,
Haoran Sun$^{1,3}$,
Hedan Yang$^{1,3}$,
Jing Long$^{1,3}$, 
Junwu Xiong$^{1}$,
Luqiao Wang$^{1,5}$,
Mingxi Luo$^{1}$, 
Qiming Yang$^{1}$,
Shuai Di$^{1}$,
Song Wang$^{1,5}$,
Tianyun Zhao$^{1,2}$,
Wanting Xu$^{1}$,
Wen Huang$^{1,2}$, 
Xiaodong Bai$^{1}$,
Xiaomeng Tian$^{1,5}$,
Xiaolong Xiang$^{1,5}$,
Yicheng Gong$^{1}$,
Yongjian Guo$^{1,2}$, 
Yucheng Guo$^{1}$, 
Yuzhen Li$^{1}$,
Yunxuan Ma$^{1,3}$, 
Yu Wei$^{1,6}$,
Zhong Guan$^{1,4}$,
Zhen Sun$^{1,5}$\\

$^*$First-Name in Alphabetical Order\\








\textbf{Affiliation}

$^1$AI Infra Team at JDT

$^2$Tsinghua University

$^3$Peking University

$^4$Tianjin University

$^5$Beihang University

$^6$University of Science and Technology of China



\bibliographystyle{unsrt}
\bibliography{main}

\clearpage



\end{document}